\documentclass[sigconf]{acmart}
\makeatletter
\def\@ACM@checkaffil{
    \if@ACM@instpresent\else
    \ClassWarningNoLine{\@classname}{No institution present for an affiliation}%
    \fi
    \if@ACM@citypresent\else
    \ClassWarningNoLine{\@classname}{No city present for an affiliation}%
    \fi
    \if@ACM@countrypresent\else
        \ClassWarningNoLine{\@classname}{No country present for an affiliation}%
    \fi
}
\makeatother

\usepackage{graphicx}
\usepackage{caption}
\usepackage{subcaption}
\usepackage{booktabs}
\usepackage{bbm}
\usepackage{numprint}
\usepackage{wrapfig}
\usepackage{siunitx}

\graphicspath{./img/}

\newcommand{\D}{\mathcal{D}}
\newcommand{\E}{\mathbb{E}}
\newcommand{\dauc}{\Delta \textrm{AUC}}

\AtBeginDocument{%
  \providecommand\BibTeX{{%
    \normalfont B\kern-0.5em{\scshape i\kern-0.25em b}\kern-0.8em\TeX}}}

\copyrightyear{2023}
\acmYear{2023}
\setcopyright{acmlicensed}\acmConference[FAccT '23]{2023 ACM Conference on Fairness, Accountability, and Transparency}{June 12--15, 2023}{Chicago, IL, USA}
\acmBooktitle{2023 ACM Conference on Fairness, Accountability, and Transparency (FAccT '23), June 12--15, 2023, Chicago, IL, USA} \acmPrice{15.00}
\acmDOI{10.1145/3593013.3594107}
\acmISBN{979-8-4007-0192-4/23/06}

\begin{document}
\sloppy
\setlength{\abovedisplayskip}{1pt}
\setlength{\belowdisplayskip}{1pt}
\title[Cross-Institutional Transfer Learning for Educational Models]{Cross-Institutional Transfer Learning for Educational Models: \\ Implications for Model Performance, Fairness, and Equity}

\author{Josh Gardner}
\email{jpgard@cs.washington.edu}
\orcid{0000-0002-4998-5918}
\affiliation{%
  \institution{University of Washington}
}

\author{Renzhe Yu}
\email{renzheyu@tc.columbia.edu}
\orcid{0000-0002-2375-3537}
\affiliation{%
  \institution{Teachers College \& Data Science Institute, Columbia University}
}

\author{Quan Nguyen}
\email{quan.nguyen@ubc.ca}
\orcid{0000-0001-8937-5121}
\affiliation{%
  \institution{University of British Columbia}
}

\author{Christopher Brooks}
\email{brooksch@umich.edu}
\orcid{0000-0003-0875-0204}
\affiliation{%
  \institution{School of Information, University of Michigan}
}

\author{Rene F. Kizilcec}
\email{kizilcec@cornell.edu}
\orcid{0000-0001-6283-5546}
\affiliation{
  \institution{Department of Information Science, Cornell University}
}

\renewcommand{\shortauthors}{Gardner et al.}

\begin{abstract}

Modern machine learning increasingly supports paradigms that are multi-institutional (using data from multiple institutions during training) or cross-institutional (using models from multiple institutions for inference), but the empirical effects of these paradigms are not well understood. This study investigates cross-institutional learning via an empirical case study in higher education. We propose a framework and metrics for assessing the utility and fairness of student dropout prediction models that are transferred across institutions. We examine the feasibility of cross-institutional transfer under real-world data- and model-sharing constraints, quantifying model biases for intersectional student identities, characterizing potential disparate impact due to these biases, and investigating the impact of various cross-institutional ensembling approaches on fairness and overall model performance. We perform this analysis on data representing over 200,000 enrolled students annually from four universities without sharing training data between institutions.

We find that a simple zero-shot cross-institutional transfer procedure can achieve similar performance to locally-trained models for all institutions in our study, without sacrificing model fairness. We also find that stacked ensembling provides no additional benefits to overall performance or fairness compared to either a local model or the zero-shot transfer procedure we tested. We find no evidence of a fairness-accuracy tradeoff across dozens of models and transfer schemes evaluated. Our auditing procedure also highlights the importance of intersectional fairness analysis, revealing performance disparities at the intersection of sensitive identity groups that are concealed under one-dimensional analysis.\footnote{Code to reproduce our experiments is available at \url{https://github.com/educational-technology-collective/cross-institutional-transfer-learning-facct-2023}.}\footnote{\copyright ~\shortauthors 2023. This is the author's version of the work. It is posted here for your personal use. Not for redistribution. The definitive version was published in ACM FAccT 2023, \url{https://doi.org/10.1145/3593013.3594107}}
\end{abstract}



\maketitle

\section{Introduction}

Improvements in digital infrastructure have enabled numerous applications of machine learning across domains but in decentralized organizational contexts (e.g., universities, schools, hospitals, finance, and government), the capacity to use machine learning typically depends on the availability of local infrastructure and resources. Under-resourced institutions may therefore be unable to reap the full benefits of machine learning applications despite commonly having the greatest need. Externally developed models have increasingly been adopted in these cases to address this challenge. While this highlights a potential benefit of sharing predictive models within large-scale cross-institutional collaborations, the risks and benefits of cross-institutional modeling are not well understood, particularly in terms of the potential impact on the most vulnerable populations affected by these models. 

\begin{figure}
  \begin{center}
    \includegraphics[width=\columnwidth]{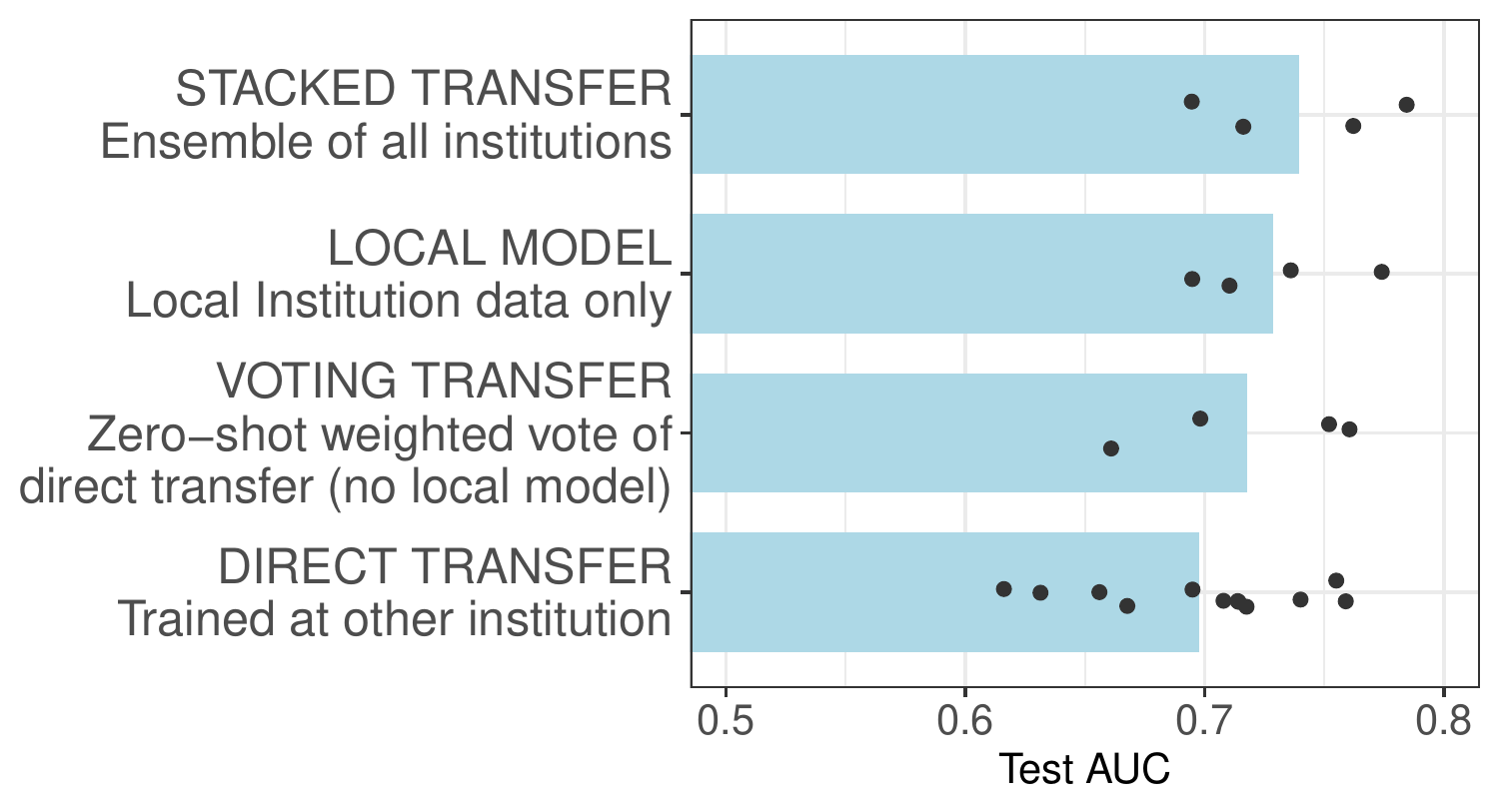}
  \end{center}
  \caption{Summary of cross-institutional transfer methods evaluated in this work (transfer methods defined in Section \ref{sec:cross-inst-transfer}). Each point represents one cross-institutional transfer trial (see Section \ref{sec:results}) with an $L_2$-regularized logistic regression model. (See Figure \ref{fig:overall-lightgbm-mlp} for results with LightGBM, MLP models.)}\label{fig:overall-l2lr}
\end{figure}

We investigate the benefits and risks of cross-institutional transfer in the context of an important and pervasive application of machine learning in education: university student dropout prediction. Every year, over one million students drop out of college in the United States, and they are 100 times more likely to default on their student loan payments than those who graduate~\cite{Kantrowitz2021}. 
This leaves young adults with a major financial burden and little to improve their job prospects with an incomplete degree
~\cite{schneider2011}. Dropout rates are especially high for students from minority groups\footnote{\url{https://nces.ed.gov/programs/raceindicators/indicator_red.asp}}, exacerbating existing inequities. 

Lowering college dropout rates is a priority for many institutions of higher education. U.S. federal regulations incentivize colleges and universities to reduce dropout by requiring them to report dropout rates, performance-based funding~\cite{hillman2014performance}, and college rankings that account for graduation rates~\cite{alsmadi2020us}. 
This has led an increasing number of colleges and universities to adopt data-driven predictive models to identify at-risk students, in order to intervene early enough to support students before they drop out. While private vendors such as Civitas, Starfish Retention, and Hobsons sell software for at-risk prediction to institutions, they do not share the technical details of underlying models. This limits the ability of practitioners and researchers to audit the performance and possible biases of these models' predictions, especially for institutions other than those where models were trained~\cite{kizilcec2021algorithmic,baker2021algorithmic}.

Concerns about algorithmic bias in education have motivated several recent studies that interrogate the fairness and ethics of predictive modeling in education~\citep{gardner2019evaluating, kizilcec2021algorithmic, lee2020evaluation, yu2021should, Yu2020}. Inter-university data partnerships, such as the Unizin Consortium\footnote{\url{https://unizin.org/}}, have also emerged to standardize data infrastructure, provide opportunities for multi-institutional model development by using data from multiple institutions during training, and even facilitate cross-institutional model sharing. However, the benefits and risks of this form of transfer learning are presently understudied, and studying cross-institutional learning in a research context can be challenging due to data privacy regulations: in most circumstances, student data cannot simply be shared between institutions or uploaded into public repositories due to federal regulations.

In this work, we seek to evaluate the implications for model \textit{performance} and \textit{fairness} of three approaches to \textit{cross-institutional transfer learning}. We conduct the first large-scale, systematic analysis of cross-institutional transfer learning in higher education. We evaluate three transfer approaches (Section~\ref{sec:cross-inst-transfer}) motivated by real-world data, collaboration models, and institutional needs in higher education. We use datasets from four U.S. universities with diverse student populations (Section~\ref{sec:data}), propose metrics to evaluate model performance and fairness in cross-institutional transfer learning (Section~\ref{sec:metrics}), conduct a comprehensive set of experiments to measure the effects of different transfer approaches on the proposed metrics in terms of (a) overall performance (Section \ref{sec:experiments-predictive-performance}) and (b) fairness measured by intersectional performance disparities (Section \ref{sec:fairness-analysis}), and evaluate fairness-performance tradeoffs (Section \ref{sec:results-tradeoff}). Our main results are summarized in Figure~\ref{fig:overall-l2lr}.
We discuss limitations and recommendations for future work in Section \ref{sec:conclusions}.

\textbf{Our contributions:} While prior research has explored prediction models based on multi-institutional educational datasets, to our knowledge, our work is the first to systematically investigate the implications of cross-institutional transfer learning for fairness and equity. This work required a year-long effort working with each institution to develop a common data schema to map their local data into this schema to enable cross-institutional learning. We contribute a novel methodology for auditing cross-institutional transfer learning, including metrics for measuring intersectional fairness of model transfer. Finally, our empirical results provide a useful benchmark for researchers and practitioners interested in cross-institutional transfer learning both within and outside of the domain of education. Our results demonstrate that (i) cross-institutional transfer is feasible even when no historical training data is present at a target institution (e.g., via direct or voting transfer); (ii) a simple zero-shot voting transfer method achieves similar performance to a local model for all institutions in our study, without requiring any local training data and at no cost to fairness; (iii) stacked ensembling provides no additional benefit over local training or zero-shot voting transfer from other institutions; and (iv) there does not exist a strict empirical tradeoff between fairness and accuracy across our broad set of models, transfer schemes, and institutions evaluated.

\section{Related Work}\label{sec:related-work}

\subsection{College Dropout Prediction}\label{sec:dropout-prediction}

Higher education institutional data are increasingly used for research and applications to predict and explain factors contributing to student dropout ~\cite{Tinto1975,Kuh2007}. These models and applications, often in the form of early warning systems, can help identify which students might be struggling and at what time in their academic journey. This information can be used to provide proactive, targeted support or interventions. Several previous studies have investigated the task of dropout prediction in higher education \cite{Aulck2019,Dekker2009,Jayaprakash2014,DelBonifro2020,Beaulac2019,Berens2019, Hutt2019, yu2021should,Bird2021,Luo2018,Chen2018,Ameri2016}. These studies are conducted across different institutional contexts, but the core learning problem they address is framed as a binary classification task, where structured features about students' educational history, demographics, or academic records in the early phase of college are extracted from administrative data to predict dropout/persistence at a later stage. Learned models are evaluated using binary classification metrics and state-of-the-art models using students' pre-college characteristics and early college records can predict whether a student will drop out within the first year with an AUC-ROC between 0.7 and 0.9 across various countries and institutions~\cite{Berens2019,Bird2021,Ameri2016,Chen2018}. These promising results have continued to motivate research into dropout prediction modeling to aid student support services and institutional policy-making. 
However, there has been little research that empirically examines cross-institutional dropout prediction. 

\subsection{Disparate Impacts and Fairness- Performance Tradeoffs in Machine Learning}\label{sec:prior-work-fairness-performance}

A great deal of prior work has explored both empirical and algorithmic approaches to ``fairness'' in machine learning, which is often concerned with how an objective function or conditional risk estimate varies across subgroups~\cite{corbett2018measure}. The current study is most closely related to previous works that explore the disparate impact of machine learning methods, and the interaction between fairness and predictive performance in machine learning.

Recent work finds that different learning techniques and contexts can result in a disproportionate impact on subgroups, even when these techniques are not explicitly targeted toward subgroups in any way, and even when they improve some average measure(s) of performance. Disproportionate impacts have been demonstrated in the use of differential privacy~\cite{bagdasaryan2019differential}, model compression~\cite{hooker2020characterising}, model simplification~\cite{kleinberg2019simplicity}, selective classification~\cite{jones2020selective}, synthetic data generation~\cite{cheng2021can}, under the presence of feature noise~\cite{khani2020feature}, and in repeated (i.e. multi-round) loss minimization~\cite{hashimoto2018fairness}.

There have been some formal analysis of potential tradeoffs between fairness and various measures of model performance~\cite{cummings2019compatibility, feldman2015certifying, yu2021should}, but our theoretical understanding remains limited in many areas. Empirically, however, there is some evidence that there are \emph{no} strict tradeoffs between model performance and fairness under certain conditions. For example, \citet{rodolfa2021empirical} finds that fairness-enhancing interventions across policy programs in education, mental health, criminal justice, and housing safety improved fairness with negligible effects on model accuracy. In education, a recent study found no evidence of a strict tradeoff between model performance and fairness predicting dropout from massive open online courses \citet{gardner2019evaluating}. However, the degree to which these findings apply to university student dropout or retention models is unknown.

Research on algorithmic fairness in education has mostly investigated whether supervised learning models trained on the entire student population generate systematically biased predictions of individual outcomes such as correct answers\cite{Doroudi2019}, course grades \cite{Yu2020}, university \cite{yu2021should} and MOOC dropout \cite{gardner2019evaluating}, learned representations of student writing \cite{arthurs2020, loukina2019}, and graduation \cite{Hutt2019, Kung2020}; it has also explored algorithms for at-risk prediction under fairness constraints \cite{Hu2020}. 
Overall, this area of research is nascent and in need of systematic frameworks specific to educational contexts to map an agenda for future research. It does suggest, however, that analyses of novel learning paradigms (such as cross-institutional transfer) should include thorough auditing for fairness, a goal of the current study.

\subsection{Cross-Institutional Learning}\label{sec:cross-inst-modeling}

We introduce the term \textit{cross-institutional} learning to describe the context where data is partitioned across a set of institutions by observation (i.e., each institution has a set of records with the same features, but collectively training on all institutions' centralized data is not possible). Prior work related to cross-institutional modeling has used a variety of different monikers, including ``cross-silo'' \cite{kairouz2019advances}, ``horizontal''  \cite{yang2019federated}, ``collaborative''  \cite{sheller2020federated}, and transfer learning. 

Recent advances in both tooling and theory have, in principle, enabled improved access to cross-institutional training. In the past five years, several usable open-source frameworks for distributed, decentralized machine learning\footnote{e.g. TensorFlow Federated \url{https://www.tensorflow.org/federated}, IBM Federated Learning \url{https://ibmfl.mybluemix.net/}} as well as frameworks for privacy-preserving learning\footnote{TensorFlow Privacy \url{https://github.com/tensorflow/privacy}, Opacus \url{https://opacus.ai/}} have emerged. Theoretical advances, such as various approaches to differentially-private model training (DP-SGD \cite{abadi2016deep}, PATE \cite{papernot2018scalable}) have provided provable guarantees regarding non-identifiability of model training examples, which may reduce privacy concerns related to cross-institutional collaboration and further pave the way for cross-institutional training in practice. The development and refinement of techniques such as federated learning and secure multi-party computation or homomorphic encryption has also enabled distributed training with improved privacy and security \cite{kairouz2019advances}.

Prior work has also investigated the related theoretical problem of learning fair models when the test distribution differs from the training distribution, as may be the case in cross-institutional learning. For example, \citet{cotter2019training} studies the use of constrained optimization to improve the satisfaction of a fairness constraint on a held-out dataset with a possibly different distribution from the training set, \citet{Coston2019} evaluates mitigating unfairness on a target domain due to covariate shift when sensitive attributes are unknown, and \citet{Singh2021} explicitly investigates fairness under distribution shift. Algorithms which are ``fairness-preserving'' have been shown to be sensitive to variations in random train-test splits \cite{friedler2019comparative}, suggesting that the fairness properties of models developed using such algorithms are brittle across distributions.

Applied research on cross-institutional learning has been somewhat limited.
In the medical domain, cross-institutional learning has been used for brain tumor segmentation \cite{sheller2018multi,sheller2020federated}, diabetic retinopathy diagnosis, and mammography screening \cite{chang2018distributed}. For example, in the context of a tumor segmentation model, direct transfer leads to average performance degradation at varying levels for 9 of 10 institutions evaluated, while collaborative learning improves performance and performs similarly to data-sharing, depending on the approach used~\cite{sheller2020federated}
and can be similar to the performance of centralized models in simulated settings \cite{matschinske2021featurecloud}. 
\citet{chang2018distributed} evaluates several transfer scenarios (local, ensembling via prediction averaging, single weight transfer, cyclical weight transfer) and finds that ensembling and weight transfer both outperformed local models in terms of validation and testing accuracy. 
\citet{pessach2021fairness} looks into the task of collaboratively training fair models across institutions through a preprocessing mechanism which leads to fairness improvement. However, the impact of the intervention on individual institutions' models and institutional subpopulations is unclear.

In the domain of education, there have been some research efforts that formulate and empirically examine the issue of model transferability across instructional, institutional, and even societal contexts~\cite{Jayaprakash2014,Ocumpaugh2014,Li2021,Ding2019,Lagus2018,Boyer2015}. For example, \citet{Ocumpaugh2014} found that models detecting students' affective states in tutoring systems do not transfer well when trained on one student population and tested on another, especially when rural students are the target (test) population. Similarly, \citet{Li2021} investigated whether academic achievement prediction models trained on U.S. samples can generalize to other national contexts, and found that the performance drops significantly for less developed countries. Most closely related to the current study, \citet{Jayaprakash2014} trained an early alert system of academically at-risk students at a liberal arts college and applied the model to four partner institutions with different institutional profiles. They found that the predictive performance declined but was still practically useful -- the recall of at-risk students only dropped from 85\% at the source institution to between 61\% to 84\% (an average of 75\%) at the target institutions. These efforts show the promise of cross-institutional educational models especially when low-resourced institutions cannot afford to develop their own model, but still require more research to ensure that performance does not degrade across institutions, harming vulnerable students.

\section{Present Study and Research Questions}\label{sec:rqs}

Under the Family Educational Rights and Privacy Act (FERPA), U.S. higher education institutions are required to maintain records about students and enrollments for purposes of external reporting (e.g., to federal educational authorities) and internal improvement. Local student information systems (SIS) are widely used to manage these records and can facilitate the identification of students who are at risk of failing classes, not graduating on time, or dropping out (see Section~\ref{sec:dropout-prediction}). Due to shared reporting responsibilities, common operational routines, and similar software tools, institutions tend to have many overlapping features in their SIS data (e.g., students' course enrollments and demographic characteristics).

Our study leverages this commonality across four universities in order to explore the impacts of cross-institutional educational modeling. As discussed in Section~\ref{sec:cross-inst-modeling} above, the limited prior work on cross-institutional transfer learning has suggested that direct transfer of learned models across institutions tends to degrade performance, and that only specialized weight-sharing strategies allow institutions to realize performance gains from transfer learning. In addition, little research has evaluated the fairness implications for such transfer scenarios. 
Building on these previous insights, our research addresses the following three research questions within the domain of dropout modeling in higher education:

\textbf{RQ1.} How does cross-institutional transfer (direct, voting, and stacked) affect performance relative to a local model?

\textbf{RQ2.} How does cross-institutional transfer affect the (intersectional) fairness of the resulting model?

\textbf{RQ3.} Is there a tradeoff between model performance and fairness under cross-institutional transfer?

\section{Methods}\label{sec:methods}

\subsection{Data and Preprocessing}\label{sec:data}

We use (de-identified) data from four public universities in the United States. All data is for first-time, first-year students in four-year bachelor's programs. Our dataset represents a wide range of enrollment sizes, demographic compositions, and first-year retention rates as summarized in Table \ref{tab:summary}.

\begin{table*}[]
    \centering
    \rowcolors{2}{white}{gray!25}
    \begin{tabular}{p{1.2cm}p{1cm}p{0.8cm}p{0.7cm}p{1cm}p{0.8cm}p{0.8cm}p{1cm}p{1cm}p{1cm}p{1.4cm}}
    \toprule
        Institution & N$_{train}$ & Female & URM & Hispanic & Asian & Black & Native Amer. & Two or more & White & First-year Retention \\ \midrule
        A & 20k &  49\% & 40\% & 26.7\% & 7.1\% & 5.1\% & 1.5\% & 5.1\% & 53.3\% & 80\% \\
        B & 1k  & 63\% &24\% & 12.6\% & 2.9\% & 5.7\% & 0.9\% & 4.3\% & 72.8\% & 57\% \\
        C & 20k & 57\% & 36\% & 33.1\% & 47.9\%& 3.2\% & 0.0\% & 0.0\% & 12.9\% & 94\% \\
        D & 30k & 55\% & 14\% & 6\%    & 11.9\%& 5.3\% & 0.0\% & 4\% & 67.5\% & 98\% \\
    \bottomrule
    \end{tabular}
    \caption{Summary statistics for the training dataset for each institution showing demographic characteristics according to federal reporting requirements (URM: underrepresented racial minority; Two or more: multiple racial/ethnic groups).}
    \label{tab:summary}
\end{table*}

We study the effects of cross-institutional transfer by converting the raw student information system (SIS) data obtained from each institution into a shared schema. 
Due to restrictions on data sharing, each institution's data was preprocessed separately and then validated by a shared pipeline prior to modeling. Only the learned model weights, not the data nor any intermediate artifacts (such as gradients during training), were shared outside of each institution. 
A goal of this project is to use SIS data in a form as close to its raw format as possible (i.e. minimal additional feature engineering), while also retaining the maximum number of viable features for experiments. In practice, this required balancing (i) removing features when insufficient data was available for all institutions or operationalization of variables was irreconcilable, and (ii) identifying ways to map related but non-identical features at each institution into a common semantic space. The process of defining a shared schema and processing the raw exports of each institution's SIS required domain expertise as well as familiarity with each institutional dataset. 

The full schema produced by all institutions for our analysis is described in Table~\ref{tab:schema}. Each row in the dataset represents a student enrolled in the fall term. The features describe students' academic history, demographics, current course load and course topics, and future plans (e.g., majors and minors). While the classification of gender as binary and the specific ethnic and racial groups raises concerns, we rely on the student categories used by the institutions themselves, which are shaped by federal reporting requirements. We provide further details on the schema in Section~\ref{sec:schema}. We release our code to validate cross-institutional datasets for conformity to this schema, and to replicate experiments using these features.\footnote{\url{https://github.com/educational-technology-collective/cross-institutional-transfer-learning-facct-2023}}

\subsection{Task}

Our target prediction is first-year \textit{retention}: for each student who enters an institution for the first time in the fall, we predict a binary indicator for whether that student will enroll at the same institution the following fall. This target matches the National Student Clearinghouse's definition \footnote{\url{https://nscresearchcenter.org/persistence-retention/}}, and is widely used both in research (Section~\ref{sec:dropout-prediction}) and practice in education.

We embrace the data constraints faced by educational institutions, which can limit the applicability of some previously proposed techniques for transfer learning. Federal regulation to protect student data privacy (FERPA) creates challenges for data sharing: costs associated with determining whether data may be shared, and then facilitating the sharing may be intractable for many institutions. We, therefore, do not consider techniques, such as federated learning, which require collaborative training in any form. For similar reasons, we do not consider approaches that require data sharing, for example, for training a centralized model on a joint dataset. Instead, we evaluate the realistic setting where each institution can only share \textit{model weights} and only a single round of cross-institutional weight sharing is possible. 
In our experiments, the barriers between institutional datasets are real, as are the challenges of cross-institutional transfer. Contrast this with prior work on cross-institutional transfer or fairness under domain shift discussed in Section~\ref{sec:cross-inst-modeling}, which frequently simulated different ``institutions'' or ``domains'' by synthetically partitioning a single dataset. 

\subsection{Three Approaches to Cross-Institutional Transfer}\label{sec:cross-inst-transfer}

Institutions in almost every domain, and particularly in higher education, differ in (i) data capacity, and (ii) modeling capacity. As a result, institutions vary in how they may be able to develop or utilize cross-institutional models. We define three distinct approaches to cross-institutional transfer, intending to cover three common contexts where institutions may seek to utilize cross-institutional models. We term these \textit{direct transfer}, \textit{voting transfer}, and \textit{stacked transfer}. As a baseline for comparison, we consider a \textit{local} model trained at the same institution where it is tested.

We consider a dataset $\D_k \coloneqq X_k, Y_k = (x_i, y_i)_{i=1}^n \sim \mathcal{P}_k$ of i.i.d. observations drawn from distribution $\mathcal{P}_k$, where $|x_i|=d$ features are present and $k$ represents an institution of interest. Denote $f(\theta, \cdot)$ as a model with parameters $\theta$, where $f(\theta, x_j) = \mathbb{P}(y_j = 1 | x_j)$ is the model's predicted probability that $x_j$ has label 1, noting that $j$ indicates a potentially different institution of interest, and thus $x_j$ may come from a distribution $\mathcal{P}_j$ which is different from $\mathcal{P}_k$. Denote the parameters estimated by training $f$ on $\D_k$ as $\hat{\theta}(\D_k)$. Denote the loss (for some general loss function) of a model trained on distribution $k$ and evaluated on distribution $\ell$  as $\mathcal{L} \Big( f(\hat{\theta}(\D_k), \tilde{X}_\ell), \tilde{Y}_\ell \Big)$, where $\tilde{\mathcal{D}_k} \coloneqq \tilde{X}_k, \tilde{Y}_k$ indicates the \textit{test} split from a distribution $k$. We use $\mathcal{I}$ to refer to the set of all institutions.

We define each of the transfer learning approaches used in our experiments as follows:

\textbf{Local:} A local model is one trained on the same institution from which it is evaluated. That is, for institution $k$, the performance of a local model is defined by $\mathcal{L} \Big( f(\hat{\theta}(\D_k), \tilde{X}_k), \tilde{Y}_k \Big)$.

\textbf{Direct Transfer:} A direct transfer scenario is one where a model is to be deployed to an institution different from its testing institution. That is, for institutions $k$, $j$, the direct transfer model performance is measured by $\mathcal{L} \Big( f(\hat{\theta}(\D_k), \tilde{X}_j), \tilde{Y}_j \Big)$. We refer to $k$ as the \textit{source} institution and $j$ as the \textit{target} institution, following the domain transfer literature. A single trained source model can be evaluated via direct transfer on several target institutions.

\textbf{Voting Transfer:} This training paradigm uses a form of averaging to combine the results of models (``voters'') trained on disjoint distributions. In our experiments, \textit{none of the voters are trained on the target institution}, which mimics the case where an institution without any historical training data uses a set of models from other institutions in a zero-shot scenario. The model under the voting transfer paradigm for target institution $i$ is defined by $\frac{1}{c} \sum_{i' \in \mathcal{I} \setminus i } f(\hat{\theta}(\D_{i'}), \cdot)$, where $c$ is the normalizing constant $|\mathcal{I}| - 1$. Note that this model does not use majority voting, but instead uses ``soft voting,'' where the predicted probabilities (not the decisions) of each model are aggregated with equal weight. This allows for the confidence of each model to be taken into account in the aggregation.\footnote{The choice of equal weighting is by convention; in practice, any combination of weights on the $(\mathcal{I} - 1)$-simplex could be used to aggregate the predictions of the voters. This weight vector could also be tuned on the target institution in a non-zero-shot formulation.}

\textbf{Stacked Transfer:} This training paradigm uses \textit{stacked generalization} \cite{wolpert1992stacked, ting1997stacked} to combine the predictions of models trained on all available institutions with the training data of the source institution. This is achieved by concatenating, for each input $x_i$, the predictions of each classifier $f(\hat{\theta}(X_j), x_i)$, to the input features, and learning a classifier from this concatenated data matrix. Formally, for institution $\ell$, if we define 

\begin{equation}
    \check{x} = [x_1, ..., x_d; f(\hat{\theta}(\D_i, x); f(\hat{\theta}(\D_j, x); f(\hat{\theta}(\D_k, x) ]\label{eqn:stacked-data-matrix}
\end{equation}
where $[ \cdot; \cdot]$ indicates column-wise concatenation, then the stacked estimator is $f(\hat{\theta}(\check{X}))$.

Because the final two forms of transfer (voting transfer, stacked transfer) are both methods for ensembling, we refer to these two methods collectively as ensemble models. 

\subsection{Metrics}\label{sec:metrics}

Metrics for evaluating various aspects of model fairness have been proposed in prior work (see Section~\ref{sec:related-work}). However, many of these metrics are based on the implicit or explicit assumption that one outcome is \textit{advantageous} or \textit{favorable}, and that a ``fair'' model can ensure some form of equity with respect to the model's predictions in placing members of sensitive subgroups into the favorable outcome class. This is often tied to contexts in which the model's predictions may be explicitly tied to some form of decision (e.g., granting a loan). Our task differs from these contexts, because the model is not explicitly tied to a decision but instead provides a prediction that might be used to assist a student, but is only useful when correct---neither predictive outcome is considered inherently ``advantageous.'' In our application, the goal of the model is to obtain equal \textit{predictive performance} for all subgroups, regardless of the true or predicted outcome. We call this \textit{equitable predictive performance}. This makes many existing fairness metrics, such as demographic parity, which assume the presence of an advantageous outcome to which we would like to equitably assign predictions, not applicable to the dropout prediction task.  There are many tasks where equitable predictive performance is desired, such as machine translation, where consistent performance is desired across dialects or languages despite differing availability of training data \cite{ahia2021low}, and image classification with respect to skin tone or other attributes \cite{buolamwini2018gender}.

Educational dropout data tends to be highly skewed by label, because in many institutional contexts, the majority of students do not drop out. This is the case in our data as well: the retention rate in our universities varies from 56\% (Institution B) to 98\% (Institution D; see Table \ref{tab:summary}). As a result, metrics such as average accuracy will tend to be biased toward majority-class predictors and will be uninformative for the small but critical subset of students who drop out.

\textbf{Area Under the Receiver Operating Characteristic Curve (AUC):} Due to our goal of equitable prediction and the significant label imbalance in our datasets, our experiments use metrics based on the Area Under the Receiver Operating Characteristic Curve (AUC), formally defined as: 
\begin{equation}
    AUC(f(\theta)) = \int_{0}^1 \textrm{TPR}(\textrm{FPR}(f_t(\theta)))dt\label{eqn:auc}
\end{equation}
where $t$ indicates a prediction threshold applied to the predictions of the model (i.e. using the decision rule $f_t(\theta, x) = \mathbbm{1} \big( f(\theta, x) \geq t\big)$), and TPR, FPR are the true positive rate and false positive rates, respectively. AUC scores are constrained to $[0, 1]$, with a random predictor achieving an AUC of 0.5. In all metrics and experiments, we compute the AUC on the test dataset, following the splitting process described in Section \ref{sec:supp-data-splitting}.

AUC is a well-studied metric of predictive performance \cite{bradley1997use, hanley1982meaning, japkowicz2011evaluating}, and has the straightforward interpretation as the probability that a randomly-selected positive example has a higher predicted probability of being positive than a randomly-selected negative example. This means that the positive and negative classes are equally weighted in computing AUC, and that AUC is less rewarding to, for example, majority-class predictors than metrics such as accuracy or cross-entropy loss. We compute standard errors for AUC values according to the procedure described in \citep{hanley1982meaning, fogarty2005case}. We provide details on computing these standard errors in Section \ref{section:seauc}.

\textbf{AUC Gap:} To measure fairness across subgroups, we define a metric that accounts for the disparities in predictive performance across a set of arbitrarily many (possibly-overlapping) subgroups $\mathcal{G}$. We define the AUC Gap as:
\begin{equation}
    \textrm{max}_{g,g' \in \mathcal{G}} ~~ | \E_{\D_k}[f(\theta(\D_{k,g})] - \E_{\D_k}[f(\theta(\D_{k,g'})] |\label{eqn:auc-gap}
\end{equation}
where $\D_{k,g}$ and $\D_{k,g'}$ indicate the subset of the data in group $g$ and $g'$, respectively. Thus, AUC Gap measures the largest difference between subpopulation AUCs, and is a measure of the worst-case performance gap between a set of subpopulations. AUC Gap is our primary measure of equitable predictive performance, because it quantifies the largest disparity in predictive performance over subgroups.

\textbf{$\Delta$AUC:} We define $\Delta$AUC to measure changes in predictive performance or fairness under cross-institutional transfer. We define the change in AUC between two transfer contexts for a fixed model $f$, as $\Delta AUC(T, T') =  AUC(T) - AUC(T')$ where, somewhat abusing notation, we use $AUC(T)$ to refer to the AUC of a model trained using a transfer scheme $T$. This allows us to compare, for example, how (overall or subgroup) AUC values are affected by transfer learning schemes. If $\dauc$ is close to zero, we can conclude that the model performs about the same in two transfer contexts; but if it is positive or negative, the model performs better/worse in context $T'$ relative to $T$. Most often in this study, we are interested in $\dauc$ relative to the local model; that is, $\dauc(\textrm{local}, \cdot)$.

\section{Experimental Setup}\label{sec:experimental-setup}

\textbf{Parameterization and Tuning of $f$:} We evaluate three forms of model transfer learning (\emph{direct transfer}, \emph{voting transfer} and \emph{stacked transfer}, described previously) plus \emph{local} models on our four institutional datasets. For each experiment, we explore three parameterizations of $f$: $(1)$ $L_2$-regularized logistic regression (L2LR); $(2)$ gradient-boosted trees (LightGBM \cite{ke2017lightgbm}); $(3)$ neural networks (multilayer perceptrons). Below, we primarily focus on L2LR in the main text due to space constraints, because (1) we observed the best performance for L2LR models across transfer schemes; (2) L2LR models are the simplest to train and tune, even for institutions with low capacity for data science and modeling; and (3) L2LR models are highly interpretable and widely used for student retention models in practice (see Section~\ref{sec:dropout-prediction}). We provide the complete results for other models in the supplementary material, including a parallel version of each figure in the main text for the other models (LightGBM, MLP); our findings with L2LR are consistent with our findings for these other models, except where explicitly noted.

Hyperparameters of each model are tuned locally via cross-validation on the source institution, and complete hyperparameter grids for each model are provided in our associated source code. For stacked models, the stacked model's hyperparameters are also tuned via cross-validation. A subset of the training data is held out and used only as a validation set for model selection (see below). The trained models are always evaluated on the test data from each institution, which is from a future academic term (see Section~\ref{sec:supp-data-splitting} for details).

As noted previously, \emph{voting transfer} is a form of zero-shot transfer and demonstrates how an institution with no training data might make use of models from other institutions. Voting transfer experiments do not use the base model from the target institution and do not require any training on the target institution: they simply use weighted majority voting to aggregate the results of each trained source model. Voting transfer allows us to evaluate the zero-shot transfer of models to an institution with no training data available. 

\textbf{Model Selection Rules:} For the \textit{stacked transfer} experiments, we explore the use of all three functional forms mentioned above (i.e. the stacked model can be L2LR, LightGBM, or MLP). However, it is not clear how the choice of base learners used to construct the stacked ensemble affects the (accuracy, fairness) of the downstream stacked model, particularly when multiple base models are available from each institution -- a realistic scenario, as it is not uncommon for an institution to train and evaluate many models during the development phase. Furthermore, it is possible that a ``greedy'' approach to selecting the base models may have unintended consequences for performance (for example, the best-performing model at Institution A may lead to poor performance when used in an ensemble at Institution C). We therefore conduct a principled investigation comparing four strategies for selecting which trained models to include in the stacking process, as follows. First, in addition to training the cross-validated models tuned at each institution, we train $10$ additional models of each type (L2LR, LightGBM, MLP) with a fixed set of hyperparameters, varying only the $L_2$ regularization of each model over a large grid. Second, from this pool of 11 candidate models at each institution (one cross-validated model plus 10 models with various degrees of regularization), we apply one of four \emph{model selection rules} to choose which estimators are included to construct \eqref{eqn:stacked-data-matrix}. Finally, we fit the stacked model and evaluate it on the target institution (the base learners are frozen throughout this process). We explore the following model selection rules:

\begin{itemize}
    \item \textit{Best Performance}: only the model with the best validation AUC at each institution is used.
    \item \textit{Best Fairness}: only the model with the best validation AUC Gap (Equation \eqref{eqn:auc-gap}) is used.
    \item \textit{Same Family}: the cross-validated models of the same functional form are used (e.g. LightGBM).
    \item \textit{Kitchen Sink}: all 11 models are used.
\end{itemize}

Due to space constraints, our results in the main text reflect the \textit{best performance} selection rule, as this is the approach most commonly used in practice. We provide additional results for all model selection rules in Section \ref{sec:selection-rules-results} (e.g. Figure \ref{fig:test-auc-all-rules}).

\section{Results}\label{sec:results}

\subsection{Overall Predictive Performance of Cross-Institutional Transfer Models}\label{sec:experiments-predictive-performance}

\begin{figure*}
    \centering
    \includegraphics[width = \textwidth]{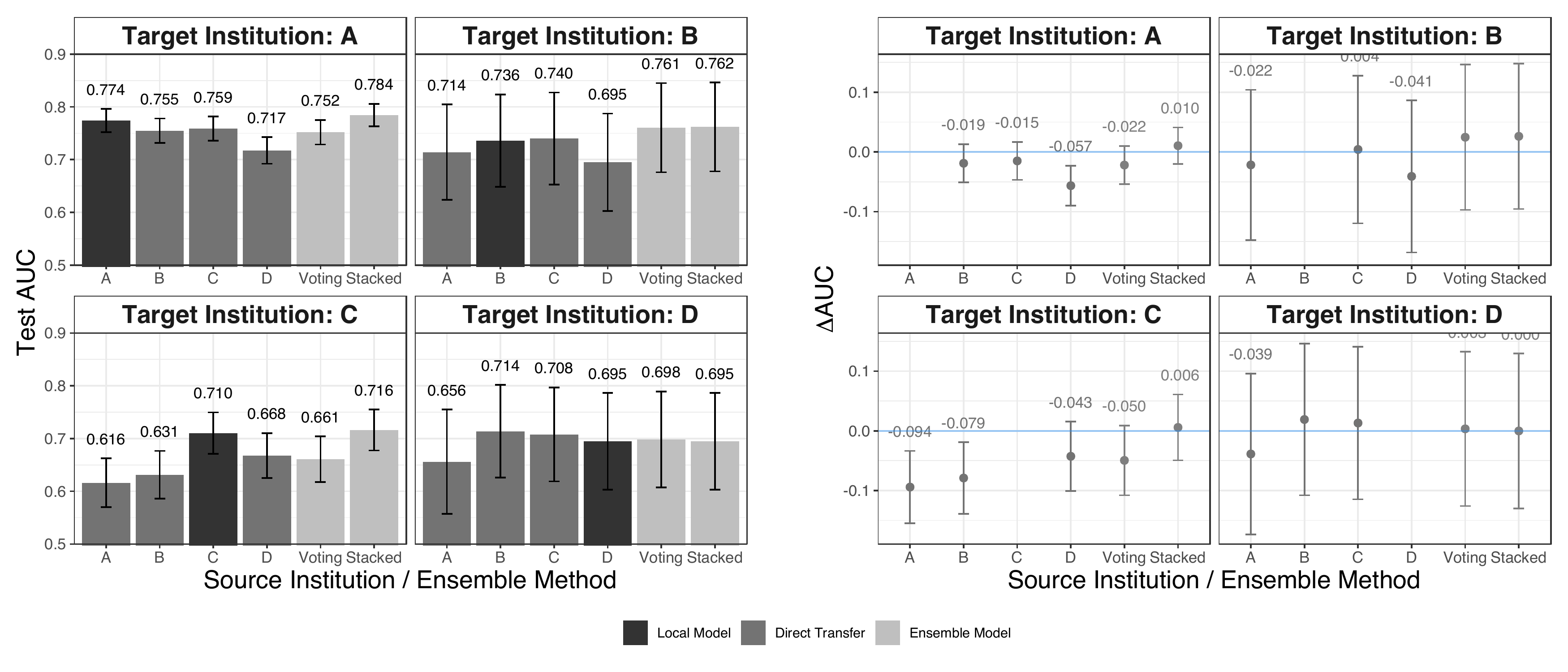}
    \caption{(a) Left: Predictive performance on test data for various transfer schemes evaluated. (b) Right: $\Delta$AUC values for various transfer models evaluated with the \emph{local} model reference line. 95\% confidence intervals shown for both figures; text displays values for point estimates. (See also Figure \ref{fig:auc-dauc-lightgbm-mlp} and Table \ref{tab:dauc-results-detail}).}
    \label{fig:auc-dauc-l2lr}
\end{figure*}

In RQ1 we are concerned with measuring model performance across three different transfer schemes (direct, voting, and stacked).
Figure~\ref{fig:auc-dauc-l2lr}a shows the AUC for models using each transfer scheme, as well as the performance of direct transfer of models from a given source institution to a target institution, for each of the four institutions in our study.
$\dauc$ values for each transfer scenario are shown in Figure \ref{fig:auc-dauc-l2lr}b.

\textbf{Direct Transfer Models:} Our results in Figure \ref{fig:auc-dauc-l2lr} show that direct transfer has inconsistent performance: for two of four institutions (B and D), all direct transfer models achieve indistinguishable performance from local models (as indicated by $\dauc = 0$ confidence intervals covering zero in Figure \ref{fig:auc-dauc-l2lr}b for direct transfer models); for the remaining two institutions (A and D), the results are mixed. This suggests, perhaps unsurprisingly, that direct transfer of models may sometimes achieve good overall performance (compared to a local model), but not in all cases.

\textbf{Ensemble Models (Voting Transfer and Stacked Transfer):} Figure \ref{fig:auc-dauc-l2lr} shows results with respect to voting and stacked transfer models. These results for both voting transfer and stacked transfer are consistent across all institutions in our study.

In our experiments, zero-shot voting transfer achieves similar performance to local models (as indicated by $\dauc = 0$ confidence intervals covering zero in Figure \ref{fig:auc-dauc-l2lr}b for voting transfer models). These results suggest that all institutions can obtain models with equivalent performance to a local model by performing zero-shot weighted aggregation of a set of models trained only on other institutions. 

Furthermore, stacked transfer provides no additional benefit over local models (or zero-shot models). Figure \ref{fig:auc-dauc-l2lr}a shows that stacked transfer models do not improve over either local models or voting transfer models (as indicated by overlapping confidence intervals for AUC between stacked and local/voting models in Figure \ref{fig:auc-dauc-l2lr}a). This suggests that once institutions have leveraged \emph{either} their own local training data (local model) \emph{or} other institutions' models (voting transfer), combining these information sources (via stacked model) provides no additional performance gains in our experiments.

A $z$-test confirms that $\dauc(\textrm{local}, \textrm{voting})$ and $\dauc(\textrm{local}, \textrm{stacked})$ are statistically indistinguishable from zero at $\alpha=0.05$ (all $p>0.1$), suggesting that the voting ensemble method is an effective way to reduce uncertainty over which direct transfer model should be used when no local training data is available, but that stacking provides no additional performance gains.  We provide exact $p$-values for the hypothesis test that $H_0: \dauc \neq 0$ in supplementary Table \ref{tab:dauc-results-detail}.
This suggests, in particular, that zero-shot transfer of a voting ensemble of three other institutions can achieve performance statistically indistinguishable from a locally-trained model.

\subsection{Intersectional Fairness Analysis of Cross-Institutional Transfer Models}\label{sec:fairness-analysis}

\begin{figure*}[tb]
    \centering
    \includegraphics[width=\textwidth]{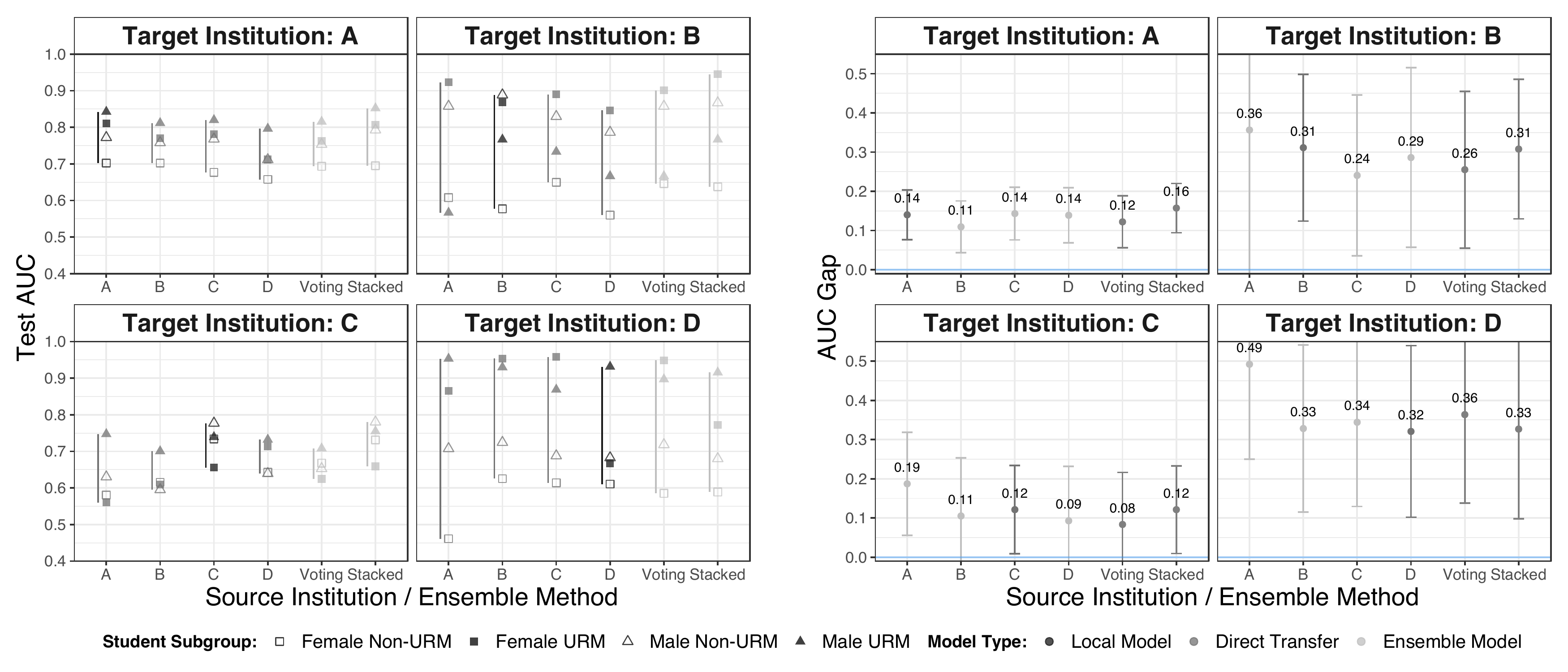}
    \caption{(a) Left: Intersectional performance over all sensitive subgroups for various transfer schemes evaluated.  (b) Right: Fairness metric \textit{AUC gap} over intersectional sensitive subgroups for various transfer schemes evaluated. 95\% confidence intervals shown; text displays values for point estimates. Note that the AUC gap (b) is the range of these values within each institution (shown as vertical bar in (a)). (See also Figures \ref{fig:aucgap-lightgbm}, \ref{fig:aucgap-mlp}, Table \ref{tab:auc-gap-results-detail}.)}
    \label{fig:aucgap-l2lr}
\end{figure*}

Our fairness analysis evaluates whether transferred models achieve equivalent predictive performance over sensitive subgroups. A substantial body of work across many disciplines and dating back several decades delineates how the \textit{intersections} of individuals' identities can contribute to disempowerment and increase vulnerability to adverse outcomes~\cite{eisestein1979combahee, crenshaw1989demarginalizing, truth1851aint}. In machine learning, however, most prior work on fairness focused on analyzing one sensitive attribute at a time (notable exceptions include \cite{foulds2019intersectional, morina2019auditing, yang2020fairness}), despite the frequent presence of multiple potentially sensitive attributes in a dataset. Therefore, we measure fairness across \textit{intersectional} subgroups via AUC Gap defined in Section~\ref{sec:metrics}. For sensitive attributes $a \in \mathcal{A}_1, a' \in \mathcal{A}_2$, we compute each metric $\mathcal{L}$ on the subset $\mathcal{D}_{a, a'} \coloneqq (x_i, y_i | A_1(x)=a, A_2(x)=a')$. 
Each metric is therefore computed as $ \mathcal{L} \big( f(\hat{\theta}(\mathcal{D}), \tilde{X}_{a,a'}), \tilde{y}_{a,a'} \big)$.
By evaluating fairness in this way, our analysis captures whether changes in AUC are distributed equally over (observable) intersecting student identities, ``focus[ing] awareness on people and experiences---hence, on social forces and dynamics---that, in monocular vision, are overlooked'' \cite{mackinnon2013intersectionality}.

We specifically use this approach to examine subgroups of students defined by sex and URM status. These represent two critical identities in the context of education with respect to which unfairness is undesirable but common in educational settings. We compute evaluation metrics for intersections of $\mathcal{A}_1 = \{ \textrm{male}, \textrm{female} \}$ for \texttt{Sex} and $\mathcal{A}_2 = \{ \textrm{Underrepresented Minority}, \textrm{Non-Underrepresented Minority} \}$ for \texttt{URM}.\footnote{Most institutions recorded more than two categories for Sex, but these tended to be "other" and "not indicated", and besides making up a small share of students, it was unclear how these responses were collected in order to interpret them correctly and consistently. We do not endorse the terminology or definition of URM; we only use it because it is consistently defined across institutions to abide by federal regulations.} The main results of our fairness analysis are shown in Figures \ref{fig:aucgap-l2lr} and \ref{fig:fairness-subgroup-data}; we also provide detailed data in supplementary Table \ref{tab:auc-gap-results-detail}. Below, we address RQ2 by separately discussing these disparity metrics under each transfer approach.

Before reporting the results, we raise two issues related to fairness, accountability, and transparency. First, the data we use is what institutions make available, and does not encapsulate all potential identities related to the sex or gender construct, and thus there is \emph{measurement bias} inherent in the data. Second, AUC (and related metrics, such as F1 score) cannot be computed unless there is at least one positively-labeled and negatively-labeled observation in each subgroup, and our data did not include any other intersectional groups where this precondition was not satisfied, and thus fairness for other measures was not evaluated, a form of \emph{algorithmic bias}.

\textbf{Local and Direct Transfer Models:} 
Figure~\ref{fig:aucgap-l2lr}a presents the four intersectional subgroup AUCs underlying the computation of each AUC gap, which is summarized in Figure~\ref{fig:aucgap-l2lr}b. The results in Figure~\ref{fig:aucgap-l2lr} show that local and direct transfer models do not differ in their performance disparities over subgroups as measured by AUC gap (Equation \eqref{eqn:auc-gap}). There is no consistent difference between the subgroup performance disparities of local vs. direct transfer models in our study, and in no case can we reject $H_0: \dauc(\textrm{local}, \textrm{direct}) = 0$.
Figure~\ref{fig:aucgap-l2lr}a also shows that a single intersectional subgroup often drives the observed performance disparities, which persist across different source models: for 17 of 24 transfer schemes evaluated, the Female Non-URM group is the lowest-AUC group for the model.

\textbf{Ensemble Models (Voting Transfer and Stacked Transfer):} In general, ensemble models achieve similar fairness relative to a local model across our experiments. Figure~\ref{fig:aucgap-l2lr-nonintersectional}b shows that, for all institutions, both voting transfer and stacked transfer achieve confidence intervals for $\dauc$ that overlap with the local model. Particularly for the voting transfer model, this is an encouraging result: It suggests that the practical benefits from the use of zero-shot voting transfer (achieving performance equivalent to a local model, \emph{without} having local training data; see Section \ref{sec:experiments-predictive-performance}) do not come at a cost to fairness, an important result. It also suggests that the improvements of voting transfer over direct transfer models do not benefit only one group; instead, the gap between the min and max AUC groups stays the same while the overall AUC improves under voting transfer (relative to direct transfer).

For 20 of 24 transfer schemes evaluated, we reject $H_0: \dauc = 0$ (see Table \ref{tab:auc-gap-results-detail} for details). This means that for most models, there is a nonzero gap between the best- and worst-performing groups for the learned model. This suggests that future work is needed to improve performance for some intersectional groups if equivalent performance across groups is possible and desirable.

\subsection{Exploratory Findings}\label{sec:results-tradeoff}

This section discusses exploratory findings from our study. Our large-scale empirical study, being the first of its kind in the domain of higher education, is uniquely positioned to provide empirical insight into several questions of interest. However, as a purely observational study with a limited set of institutions, it is our intention to clearly position our discussion of the following findings as \emph{exploratory}: our study provides initial evidence, but does not \emph{prove} there is a relationship, particularly a causal relationship, between the factors discussed here.

\textbf{No tradeoff between fairness and accuracy:} RQ3 in our study concerns whether there is a tradeoff between fairness and accuracy in cross-institutional transfer. Our results suggest that the variation in AUC Gap is epxlained by other factors (namely, institution and the transfer type), and that AUC Gap is \emph{not} associated with AUC after controlling for these factors. We explore a simple linear regression of AUC Gap on AUC, with terms for the target institution and the transfer type. We find that in the resulting linear model, the AUC term $\beta_{\textrm{AUC}}$ has $t=-0.350$ ($p=0.72719$), suggesting that we do \emph{not} have evidence to suggest that AUC and AUC Gap are associated, after controlling for the target institution and transfer type. We show a scatter plot of the data used to conduct this analysis in supplementary Figure \ref{fig:auc-vs-aucgap}, and provide details on the model, in Section \ref{sec:additional-results}.

\textbf{Impact of Intersectional Analysis:} In Section \ref{sec:marginal-vs-intersectional}, we briefly compare the findings of our intersectional analysis with a non-intersectional (``marginal'') analysis. This comparison demonstrates that marginal analyses are more likely to ignore performance disparities within subgroups, and to assign lower overall AUC Gap scores to models.

\textbf{No clear impact of regularization:} We also study whether effective regularization helps reduce subgroup performance disparities. One potential interpretation of the disparities measured by AUC Gap might be that models simply overfit to certain groups; in this case, effectively tuning the regularization parameter might reduce the degree of overfitting to certain groups. To investigate the impact of regularization, we conduct a sweep of the $L_2$ regularization for all models (L2LR, LightGBM, MLP; all contain an $L_2$ regularization term) and keep all other hyperparameters at default values. We provide the results of this study in Figure \ref{fig:lambda-study}, and give further detail on the design of these experiments in Section \ref{sec:results-regularization}. Our results suggest that there is not a clear relationship between regularization and AUC Gap. This aligns with existing work on subgroup robustness, which suggests that subgroup performance tends to improve along with the overall model (in which case regularization should be tuned to optimize the bias-variance tradeoff) \cite{gardnersubgroup}.

\section{Conclusion}\label{sec:conclusions}

This paper presents the first large-scale empirical study of model performance and fairness in cross-institutional transfer learning. We proposed a set of metrics for quantifying cross-institutional transfer performance and fairness, and applied those techniques in the context of university student dropout prediction with real-world education data. Our results show that cross-institutional transfer is possible, where even zero-shot ``voting transfer'' models achieve statistically indistinguishable performance to a local model with no change in intersectional subgroup fairness. Additionally, our results show more broadly that there is no evidence of a performance-fairness tradeoff across a wide scope of functional forms (L2LR, LightGBM, MLP), transfer schemes (direct, voting, stacked), and selection rules for the ensemble components (best performance, best fairness, same family, kitchen sink).

These findings have important implications for both researchers and practitioners. For machine learning researchers, the demonstrated success of relatively simple approaches (e.g. voting transfer) suggests that further investigation is needed to understand the conditions under which $(i)$ zero-shot transfer is effective, and $(ii)$ more sophisticated transfer learning and ensembling methods succeed or fail. For educational researchers, the results suggest that while institutional contexts matter in understanding and predicting student dropout, there exists a decent level of generalizable knowledge that can facilitate the development of portable predictive models. For practitioners, our results show that the cross-institutional learning paradigm can serve as a viable means for well-resourced institutions to support their underresourced counterparts. This is especially relevant in an era with growing availability of big data and increasing prevalence of algorithmic decision making. 

On the other hand, this study has a handful of limitations. First, our models are trained on available data from each institution, which could contain their own biases~\cite{Barocas2019,kizilcec2021algorithmic}. Treating institutional datasets as a source of truth masks complexities in how variables are coded and in how historical inequities are manifested in the data, which could affect the applicability of our conclusions. Second, our sample only includes four institutions in the United States that have the data infrastructure and capacity to make large-scale data available for research. 
This limits the generalizability of our results to other institutions with varying degrees of similarity in student populations and dropout-generating processes. The breadth of cultural perspectives in non-U.S. contexts, as well as the definition of which students are under-represented, also suggest a need for replication and extension of this work.

Informed by the limitations, there are a few lines of future work. First, further studies on cross-institutional transfer are needed, including similar studies in education and other decentralized organizations with large-scale shared electronic record-keeping systems (e.g., hospitals, local governments, financial institutions). Future work should evaluate additional transfer approaches and could include the development of algorithms designed explicitly to mitigate performance disparities. Second, cross-institutional collaborative modeling is still difficult, due to a combination of data-sharing restrictions and a lack of technical infrastructure for cross-institutional collaboration with private data. We encourage the development of better technical and theoretical frameworks for collaborative learning in the presence of strict data-sharing constraints. Finally, our work suggests that overall performance can be improved without a strict cost to fairness, providing a motivation for further improvement of general classification techniques for student retention modeling, even without explicit disparity-mitigating interventions.

\bibliographystyle{ACM-Reference-Format}
\bibliography{ref.bib}

\clearpage 

\appendix

\section{Acknowledgements}

This research was partly funded by a Google gift via the 2021 Award for Inclusion Research program. Josh Gardner was supported by a grant from Microsoft.

\section{Data}\label{sec:data-supplementary}

\subsection{Schema}\label{sec:schema}

The complete schema for the institutional data used in this study is shown in Table~\ref{tab:schema}. As discussed in Section~\ref{sec:data}, the data is extracted from each institution's student information system (SIS), and is a subset of the data available at each institution. These variables were chosen based on a combination of availability, expected predictive utility (based on prior research discussed in Section~\ref{sec:dropout-prediction} and the researchers' own experience), and consistency of coding across institutions.

\begin{table*}[]
\small
\rowcolors{2}{gray!12}{white}\rowcolors{2}{white}{gray!25}
    \begin{tabular}{p{2.5cm} p{4cm}  lccllcc}
        \toprule
        \textbf{Feature Name} & \textbf{Description} & \textbf{Type} &  & \textbf{Nullable} & \textbf{Min} & \textbf{Max} & \textbf{Student} & \textbf{Multiple Columns} \\ \midrule
        \textbf{Units} & Credit units the student enrolled in in this term & Float &  &  & 0 & 100 & & \\
        \textbf{Units Failed} & Percentage of enrolled units the student failed & Float &  &  & 0 & 100 & & \\
        \textbf{Units Incomplete} & Percentage of enrolled units the student earned an incomplete for & Float &  &  & 0 & 100 & & \\
        \textbf{Units Withdrawn} & Percentage of enrolled units the student withdrew from & Float &  &  & 0 & 100 & & \\
        \textbf{Cumulative GPA} & Cumulative GPA from all known tertiary education sources & Float &  &  & 0 & 4 & & \\
        \textbf{Units Transferred} & Total number of transferred units in this term & Integer &  & \checkmark & 0 & 100 & \checkmark & \\
        \textbf{Age} & Age at course start & Integer &  & \checkmark & 0 & 150 & \checkmark & \\
        \textbf{High School GPA} & GPA in high school & Float &  & \checkmark & 0 & 4 & \checkmark & \\
        \textbf{ACT English} & ACT English component score & Integer &  & \checkmark & 0 & 36 & \checkmark & \\
        \textbf{ACT Math} & ACT Math component score & Integer &  & \checkmark & 0 & 36 & \checkmark & \\
        \textbf{SAT Math} & SAT Math component score & Integer &  & \checkmark & 0 & 800 & \checkmark & \\
        \textbf{SAT Verbal} & SAT Verbal component score & Integer &  & \checkmark & 0 & 800 & \checkmark & \\
        \textbf{GPA Mean} & The term-level gpa average weighted by units & Float &  &  & 0 & 4 & & \\
        \textbf{GPA Stddev} & The term weighted gpa stddev & Float &  &  & 0 & inf & & \\
        \textbf{GPA $z$-score} & The weighted average z-score of the student in their courses & Float &  &  & -inf & inf & & \\
        \textbf{GPA $z$-score stddev} & The weighted stddev z-score of the student in their courses & Float &  &  & 0 & inf & & \\
        \textbf{Units Per CIP2} & Units taken by 2-digit CIP code & Float &  &  & 0 & 100 & & \checkmark \\
        \textbf{Units Per Course} & Units taken by course format & Integer &  &  & 0 & 100 & & \checkmark \\
        \textbf{Units Online} & Units taken online. & Integer &  &  & 0 & 100 &  & \\
        \textbf{Units In-Person} & Units taken in person. & Integer &  &  & 0 & 100 &  & \\
        \textbf{Modality} & Whether the student is enrolled in-person or online & Categorical &  &  &  & \checkmark & &  \\
        \textbf{Sex} & Self-declared sex & Categorical &  &  &  &  & \checkmark & \\
        \textbf{Ethnicity} & Self-declared ethnicity & Categorical &  &  &  &  & \checkmark & \\
        \textbf{URM Status} & Institutionally-assigned indicator for underrepresented minority status & Categorical &  &  &  &  & \checkmark & \\
        \textbf{Major 1 CIP Code} & 2-digit CIP code for first major & Categorical &  &  & 1 & 61 & \checkmark & \\
        \textbf{Major 2 CIP Code} & 2-digit CIP code for second major & Categorical &  &  & 1 & 61 & \checkmark \\
        \textbf{Minor 1 CIP Code} & 2-digit CIP code for first minor & Categorical &  &  & 1 & 61 & \checkmark & \\
        \textbf{Minor 2 CIP Code} & 2-digit CIP code for second minor & Categorical &  &  & 1 & 61 & \checkmark & \\
        \textbf{Year} & The current year & Integer &  &  & 2013 & 2019 & \checkmark & \\
        \textbf{Retention} & Indicator for whether student was enrolled in following fall term & Binary &  &  & 0 & 1 & \checkmark & \\ \bottomrule
    \end{tabular}
    \caption{Data schema used for this study. Each row in the resulting dataset represents a single first-year student present in the Fall academic term at an institution. ``Student-level'' features indicate those which are fixed for a given student under normal circumstances, and do not vary by term. For more detail on categorical codings and handling of nullable features, see Section~\ref{sec:schema} Note that min/max values indicate the min/max enforced by our data validation pipeline; these are not the min/max values occurring in the data (which often fall into a much smaller range).}\label{tab:schema}
\end{table*}

The ``CIP2 code'' referenced in Table \ref{tab:schema} refers to the Classification of Instructional Programs (CIP) codes defined by the Center for Educational Statistics\footnote{\url{https://nces.ed.gov/ipeds/cipcode/}}. CIP codes are written as decimal values (i.e. 14.43 is the CIP code for ``Biochemical Engineering''). We use the ``coarse'' CIP codes, which are represented by the integer value preceding the decimal (i.e. 14 represents ``Engineering''). A complete list of CIP codes may be viewed at \url{https://nces.ed.gov/ipeds/cipcode/browse}.

For features marked as ``multicolumn'' in Table~\ref{tab:schema}, several identical features are generated to represent the information for that row. For example, for ``Units Per Course Type'', we generate 62 features, where each feature indicates the number of units associated with a given CIP code (1-61, or ``Missing'' when the CIP code is not available).

The ``Retention'' variable indicates the prediction target for this study, and is an indicator for whether the student was enrolled in the following fall term, according to the university's enrollment records. Note that this is only a 1-year measure of retention; it does not measure whether the student persists to complete a degree.

Additional categorical variables are coded as follows:

\begin{itemize}
\item \textbf{Course Component:} a variable indicating which ``component'' of a course an individual record is assigned to (note that courses may sometimes also consist of multiple components, such as a lecture and a lab component). These also include multiple-component courses, which are simply the set of all combinations of the following course components: Lecture, Discussion, Lab, Seminar, Other. 
    \item \textbf{Sex:} a self-declared variable representing the students' declared sex or gender identity. The procedure for collecting this data, along with the exact allowed values, vary by institution; we include the following possible values, but recognize the limitations of such a coding: Male, Female", Not Indicated, Other.
    \item \textbf{Ethnicity:} a self-declared variable representing the students declared ethnicity. The procedure for collecting this data, along with the exact values, vary by institution. As above, we include the following possible values, which reflect the union of categories across our institutions, but recognize the limitations of this coding: Asian, Black, Hawaiian, Hispanic, Native American, Not Indicated, White, 2 or More.
    \item \textbf{URM Status:} The term `Underrepresented Minority' holds a specific institutional meaning in higher education, where it is used to refer to a category of domestic students (those with U.S. citizenship status) who hold membership in an underrepresented racial or ethnic group in the United States. We note that this is a category that is tracked and reported by almost every accredited institution in the United States. As a result, we decided to include this variable, instead of deriving a potentially more socially meaningful, but less contextual, ``underrepresented minority'' feature. This variable takes the following values: Non-Underrepresented Minority, Underrepresented Minority, International.
\end{itemize}

For variables marked as ``nullable'' in Table \ref{tab:schema}, handle them in two distinct ways: for some variables (high school GPA, ACT English, ACT Math), we drop records with those values not present (since missingness is rare for these features). For the remaining nullable features (SAT Math, SAT Verbal) we use median imputation.

We note that for \textit{all} institutions, we only use records up to the Fall 2019 term in order to avoid forecasting into the academic terms affected by the COVID-19 crisis.

\subsection{Train/Test/Validation Split}\label{sec:supp-data-splitting}

Our goal is to realistically evaluate models' ability to predict on future data. To do so, we use as training data records from all terms prior to Fall 2019 term. We reserve all records from the Fall 2019 term as validation/testing data, where these records are split evenly into test/validation.

\section{Open-Source Code Release}

Concurrent to the release of this paper, we will make our code publicly available via public GitHub repository. This includes code for data validation, model training, and evaluation, as well as other reproducibility details (software requirements, code for generating figures). 

\section{Additional Results}\label{sec:additional-results}


\subsection{Transfer Gap $\dauc$ Detailed Results}

We provide detailed experimental results for the transfer gap measure $\dauc$ in Table \ref{tab:dauc-results-detail}, and detailed experimental results for the AUC Gap fairness measure in Table \ref{tab:auc-gap-results-detail}.

\begin{table*}[]
\rowcolors{2}{white}{gray!25}
\begin{tabular}{p{0.8in} p{0.8in} S[round-mode=places, round-precision=3] S[round-mode=places, round-precision=3] l S[round-mode=places, round-precision=3]}
\toprule
\textbf{Target \newline Institution} & \textbf{Source \newline Institution(s)} & \textbf{$\dauc$}    & \textbf{$\textrm{SE}(\dauc)$} & \textbf{Transfer Type} & $p$          \\ \midrule
A                           & D-C-B                          & -0.0219996595216279   & 0.0163204086777409     & Majority Voting        & 0.177663336630349   \\
A                           & D-C-A-B                        & 0.0104071220064189    & 0.0156111665238642     & Stacked                & 0.50499826725098    \\
A                           & D                              & -0.056518158012007    & 0.0170300940400094     & Direct                     & 0.00090430213741395{*} \\
A                           & C                              & -0.0150549021375033   & 0.0161715996988594     & Direct                     & 0.351880983453849   \\
A                           & B                              & -0.0189979421086627   & 0.016256318917034      & Direct                     & 0.24254479833653    \\
B                           & D-C-A                          & 0.0245901639344261    & 0.0621333417080871     & Majority Voting        & 0.69227885080052    \\
B                           & D-C-A-B                        & 0.0261270491803278    & 0.0620566011127681     & Stacked                & 0.673740724900487   \\
B                           & D                              & -0.0409836065573771   & 0.0649766654422485     & Direct                     & 0.528208279850642   \\
B                           & C                              & 0.00409836065573765   & 0.0631122992550568     & Direct                     & 0.948223690710095   \\
B                           & A                              & -0.0217725409836066   & 0.0642307692491104     & Direct                     & 0.734629548960961   \\
C                           & D-A-B                          & -0.0495325014725696   & 0.0297720952400247     & Majority Voting        & 0.0961679195624025  \\
C                           & D-C-A-B                        & 0.00571424702608636   & 0.0281341117052904     & Stacked                & 0.839051067183201   \\
C                           & D                              & -0.0426977474763205   & 0.0295813612372278     & Direct                     & 0.148907722595508   \\
C                           & A                              & -0.0942410681038044   & 0.030926254384457      & Direct                     & 0.00230919721289636{*} \\
C                           & B                              & -0.0790448270492414   & 0.0305528003372306     & Direct                     & 0.00967720783921469{*} \\
D                           & C-A-B                          & 0.00332444771271856   & 0.0659077913192535     & Majority Voting        & 0.959771051599402   \\
D                           & D-C-A-B                        & -0.000199160462052483 & 0.0661590870339205     & Stacked                & 0.997598110480144   \\
D                           & C                              & 0.0130679903177374    & 0.0652038062086832     & Direct                     & 0.841153996313111   \\
D                           & A                              & -0.0387903299935657   & 0.0687850064471452     & Direct                     & 0.572797801583429   \\
D                           & B                              & 0.0189508839660507    & 0.064772567526943      & Direct                     & 0.769846434720799  \\ \bottomrule
\end{tabular}
\caption{Detailed results for transfer gap $\dauc$ for each transfer scheme evaluated. The final column gives the $p$-value of the hypothesis test of $\dauc \neq 0$ for the transfer scheme evaluated. $*$ indicates we reject $H_0$ at $\alpha=0.05$.}\label{tab:dauc-results-detail}
\end{table*}

\begin{table*}[]
\rowcolors{2}{white}{gray!25}
\begin{tabular}{p{0.8in} p{0.8in} S[round-mode=places, round-precision=3] S[round-mode=places, round-precision=3] l S[round-mode=places, round-precision=2]}
\toprule
\textbf{Target \newline Institution} & \textbf{Source \newline Institution(s)} & \textbf{AUC Gap} & \textbf{SE(AUC Gap)} & \textbf{Transfer Type} & \textbf{p} \\ \midrule
A & D-C-B & 0.122147108408432 & 0.0338838712693502 & Majority Voting & 0.000312303633993873{*} \\
A & D-C-A-B & 0.157333514580536 & 0.0320923000211468 & Stacked & 9.46093956341077e-07{*} \\
A & B & 0.10940552766979 & 0.0337171450071735 & Direct & 0.00117531481757619{*} \\
A & C & 0.143255245996841 & 0.0342742056757208 & Direct & 2.91918306575532e-05{*} \\
A & D & 0.138921799024389 & 0.0359921716457721 & Direct & 0.000113489993898567{*} \\
A & A & 0.140084010749891 & 0.0322570658332607 & Direct & 1.40717411216676e-05{*} \\
B & D-C-A & 0.255265567765568 & 0.102071050725772 & Majority Voting & 0.0123891595829669{*} \\
B & D-C-A-B & 0.307800043094161 & 0.0908129882330257 & Stacked & 0.000700499695884281{*} \\
B & A & 0.356410256410256 & 0.191266473105912 & Direct & 0.0624028417805434{*} \\
B & C & 0.240600086188322 & 0.104682616396358 & Direct & 0.0215403585862442 \\
B & D & 0.286104826546003 & 0.116851763491944 & Direct & 0.0143475319620179{*} \\
B & B & 0.311339009287926 & 0.0955143821178903 & Direct & 0.00111568085369003 \\
C & D-A-B & 0.0834187375572649 & 0.0677486912775264 & Majority Voting & 0.218211929157495 \\
C & D-C-A-B & 0.121194472162624 & 0.0570680544999668 & Stacked & 0.0336966378971663 \\
C & A & 0.187121457321451 & 0.0670134643994959 & Direct & 0.00523353871252754{*} \\
C & B & 0.105083255068609 & 0.0754565264267836 & Direct & 0.163730773340565 \\
C & D & 0.092819327822257 & 0.0707396295368241 & Direct & 0.189477526093772 \\
C & C & 0.121227004351214 & 0.0575149419171092 & Direct & 0.0350527921151603{*} \\
D & C-A-B & 0.363635955988274 & 0.115238014052529 & Majority Voting & 0.00160211815240661 \\
D & D-C-A-B & 0.326687906724512 & 0.116687747416188 & Stacked & 0.00511538937767855 \\
D & A & 0.492050042178838 & 0.123445420972504 & Direct & 6.72043211758981e-05{*} \\
D & B & 0.328156966935154 & 0.108867147434087 & Direct & 0.00257582920702248{*} \\
D & C & 0.344136160576659 & 0.109620474434489 & Direct & 0.00169327694144693{*} \\
D & D & 0.320741443721379 & 0.111588574552381 & Direct & 0.00404896515858058{*} \\ \bottomrule
\end{tabular}
\caption{Detailed results for fairness measure AUC Gap for each transfer scheme evaluated. The final column gives the $p$-value of the hypothesis test of $\textrm{AUC~Gap} \neq 0$ for the transfer scheme evaluated. $*$ indicates we reject $H_0$ at $\alpha=0.05$.}\label{tab:auc-gap-results-detail}
\end{table*}

\subsection{Subgroup-Specific Transfer Detail}

\begin{figure*}[]
    \centering
    \includegraphics[width=\textwidth]{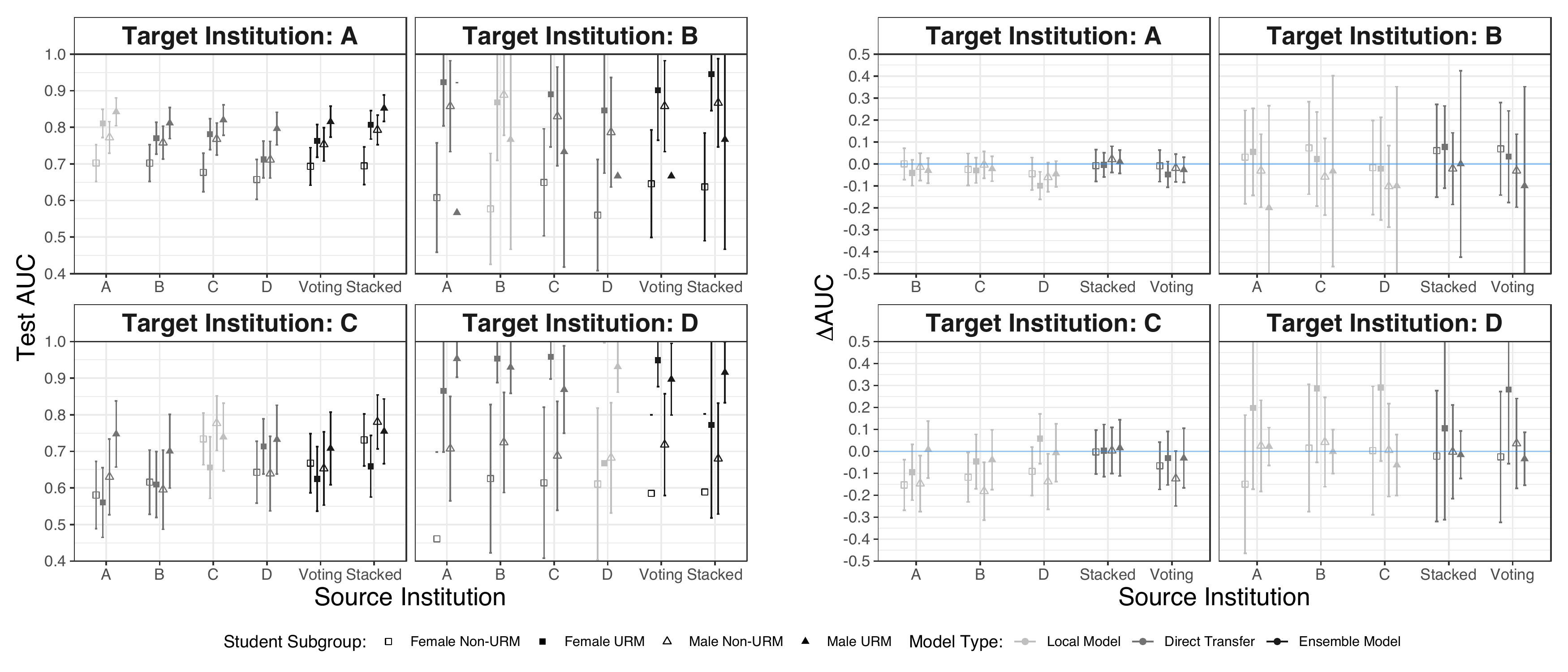}
    \caption{(a) Left: Model performance over intersectional subgroups for various transfer schemes evaluated (Male, Female, URM, Non-URM) for all institutions. (b) Right: $\Delta$AUC values over sensitive subgroups for various transfer schemes evaluated. One-SE error bars shown for both figures. Direct and local transfer models are in lexicographic order (A, B, C, D) within each subgroup.}
    \label{fig:fairness-subgroup-data}
\end{figure*}

This section gives additional results regarding intersectional model performance discussed in Section \ref{sec:fairness-analysis}. Figure \ref{fig:fairness-subgroup-data} provides the results of each model transfer scenario (local, direct, voting, stacked), organized by intersectional subgroups.

Figure~\ref{fig:fairness-subgroup-data} provides additional evidence regarding the zero-shot transfer capacity of models learned via direct transfer and voting transfer: their subgroup performance tends to be similar to the local model, with no discernable effect on performance disparities, measured by AUC gap. Most transfer schemes also have limited effect on subgroup performance relative to the local model, measured by $\dauc(\textrm{local}, \cdot)$, as shown in Figure~\ref{fig:fairness-subgroup-data}b. The exception to this is the stacked transfer model, which tends to have both improved performance relative to the local model (high $\dauc(\textrm{local}, \textrm{stacked})$) and reduced disparities between groups (low AUC gap; see Figure~\ref{fig:aucgap-l2lr}).

\subsection{Non-Intersectional Comparison for Figure \ref{fig:aucgap-l2lr}}\label{sec:marginal-vs-intersectional}

In Section~\ref{sec:fairness-analysis}, we discuss the significance of analyzing model performance disparities via intersectional groups. Here, we present evidence of the difference between the intersectional and non-intersectional (which we refer to as ``marginal'') analysis.

\begin{figure*}
    \centering
    \includegraphics[width=\textwidth]{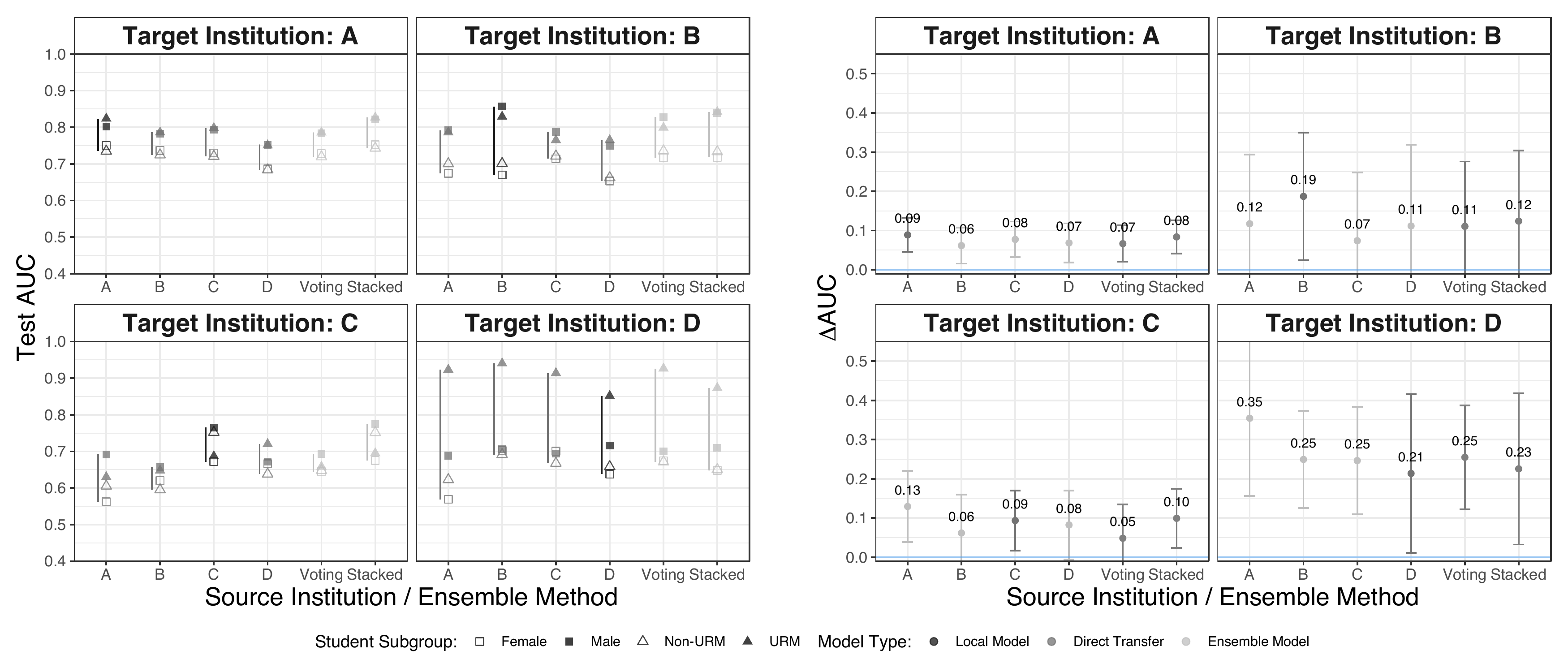}
    \caption{Non-intersectional (``marginal'') version of Figure \ref{fig:aucgap-l2lr}, where AUC is computed over marginal groups, not intersectional groups. In comparison to Figure~\ref{fig:aucgap-l2lr}, this analysis shows considerably smaller disparities. This demonstrates how marginal analyses can mask intersectional performance disparities.}
    \label{fig:aucgap-l2lr-nonintersectional}
\end{figure*}

Figure~\ref{fig:aucgap-l2lr-nonintersectional} shows an identical analysis as Figure~\ref{fig:aucgap-l2lr}, with the exception that subgroup performance is computed over marginal (non-intersectional) subgroups. Here, we can see that, for each subgroup, the performance of the model is a weighted average of the previous intersectional groups, reducing the observed performance disparities and masking larger disparities within the intersectional groups.

\subsection{Impact of Ensemble Selection Strategies}\label{sec:selection-rules-results}

\begin{figure*}
    \centering
    \includegraphics[width=\textwidth]{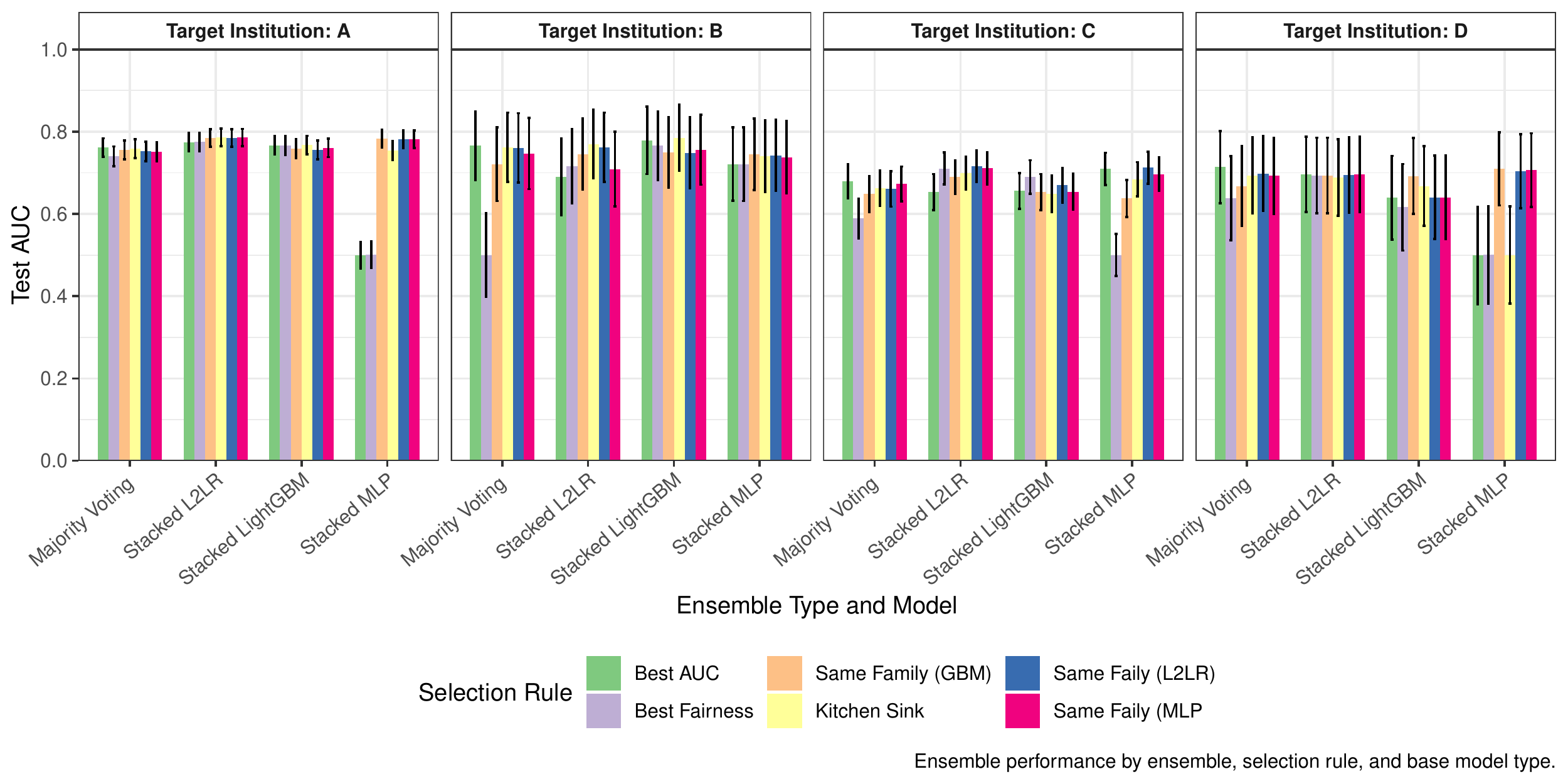}
    \caption{Test AUC by ensemble type and selection rule.}
    \label{fig:test-auc-all-rules}
\end{figure*}

We provide results comparing the impact of different model selection strategies in Figure \ref{fig:test-auc-all-rules}. 

\subsection{Impact of Functional Form for Base and Stacked Ensemble Models}

In this section, we provide additional results for LightGBM, MLP models not discussed in the main text due to space limitations.

\begin{figure*}
\centering
\begin{subfigure}{.49\textwidth}
  \centering
  \includegraphics[width=\textwidth]{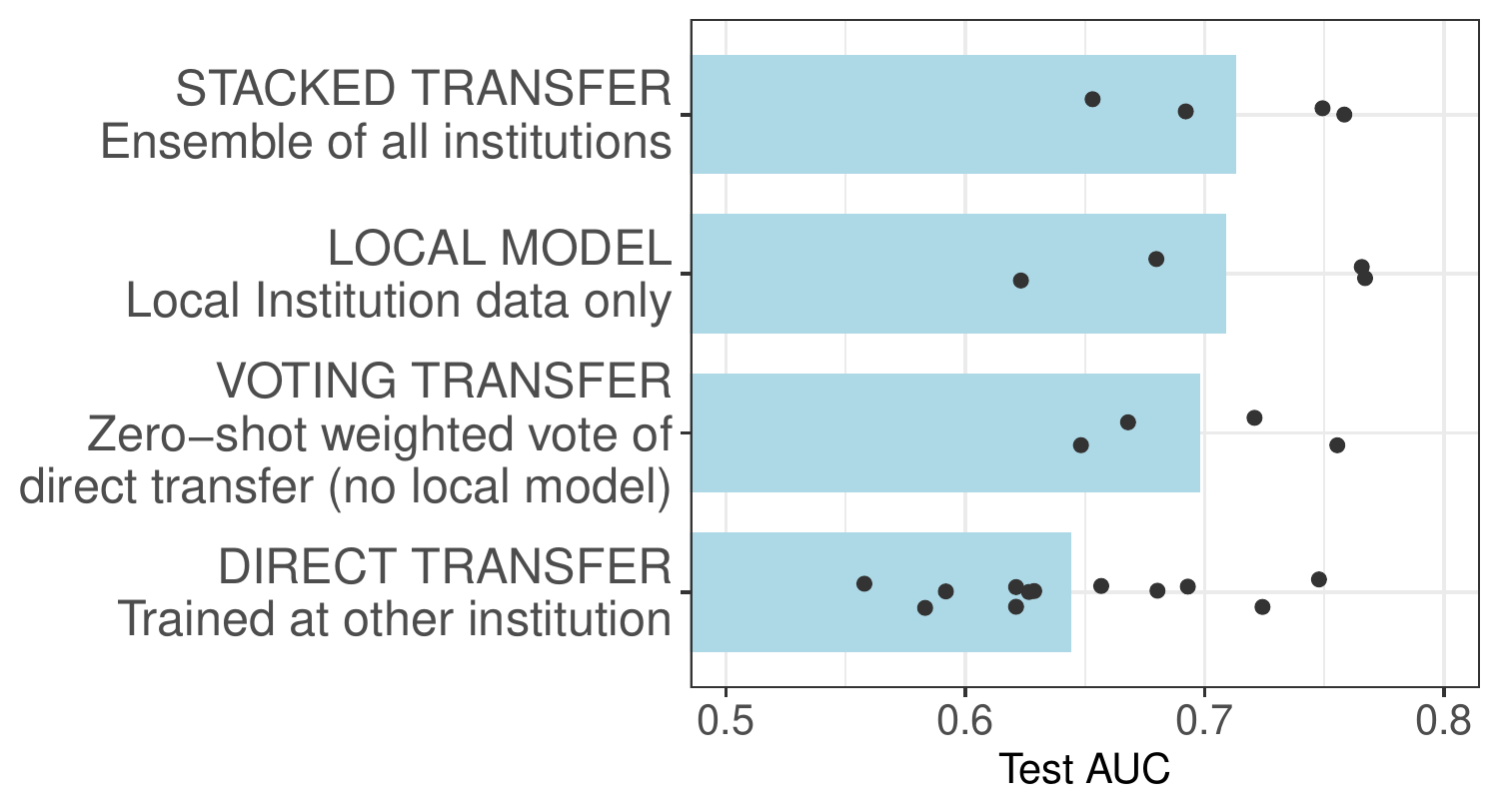}
  \caption{Overall results for LightGBM model.}
  \label{fig:overall-lightgbm}
\end{subfigure}
\begin{subfigure}{.49\textwidth}
  \centering
  \includegraphics[width=\textwidth]{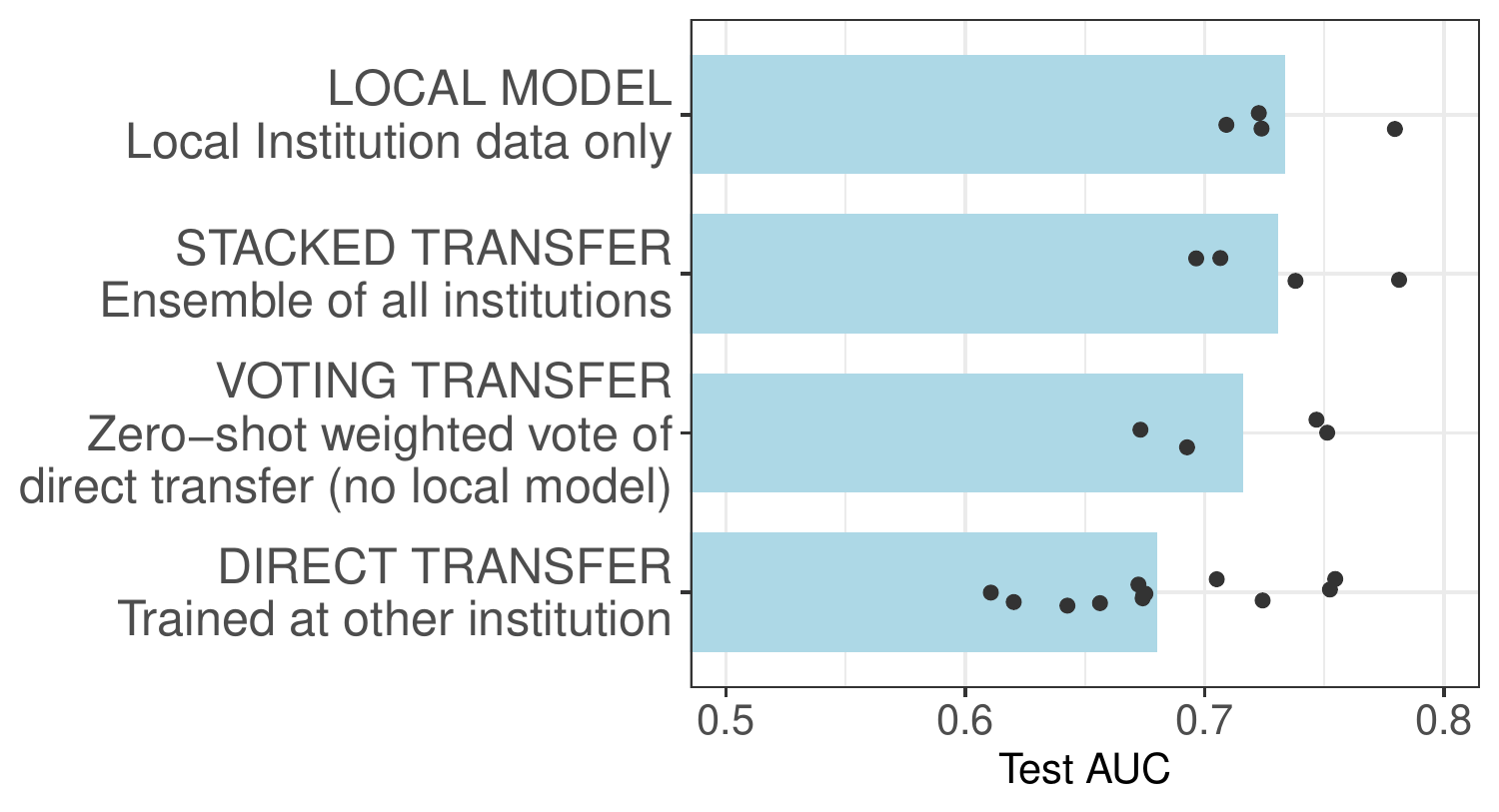}
  \caption{Overall results for MLP model.}
  \label{fig:overall-mlp}
\end{subfigure}%
\caption{Overall results for LightGBM (\ref{fig:overall-lightgbm}) and MLP (\ref{fig:overall-mlp}) models. The findings are consistent with our findings for $L_2$-regularized logistic regression shown in Figure \ref{fig:overall-l2lr}.}
\label{fig:overall-lightgbm-mlp}
\end{figure*}

\begin{figure*}
\centering
\begin{subfigure}{\textwidth}
  \centering
  \includegraphics[width=\textwidth]{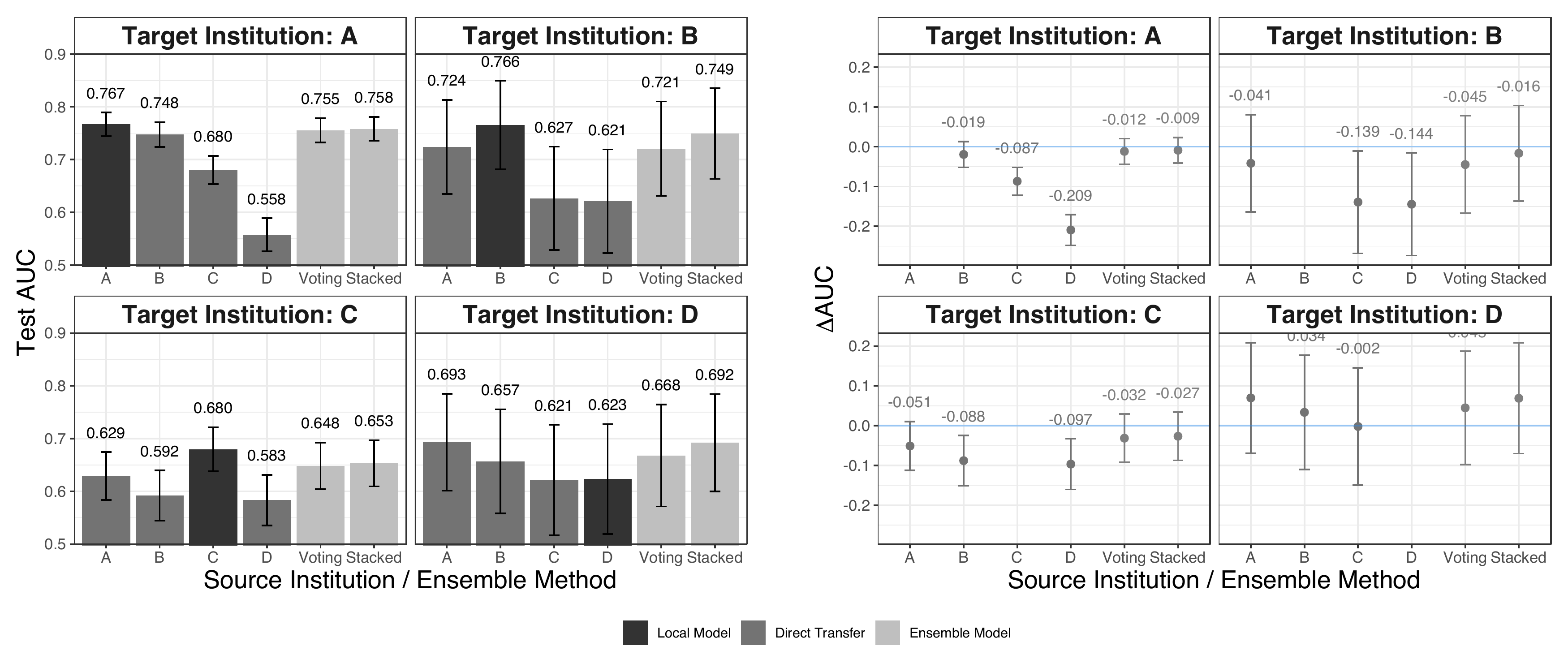}
  \caption{Results for LightGBM model.}
  \label{fig:auc-dauc-lightgbm}
\end{subfigure}
\begin{subfigure}{\textwidth}
  \centering
  \includegraphics[width=\textwidth]{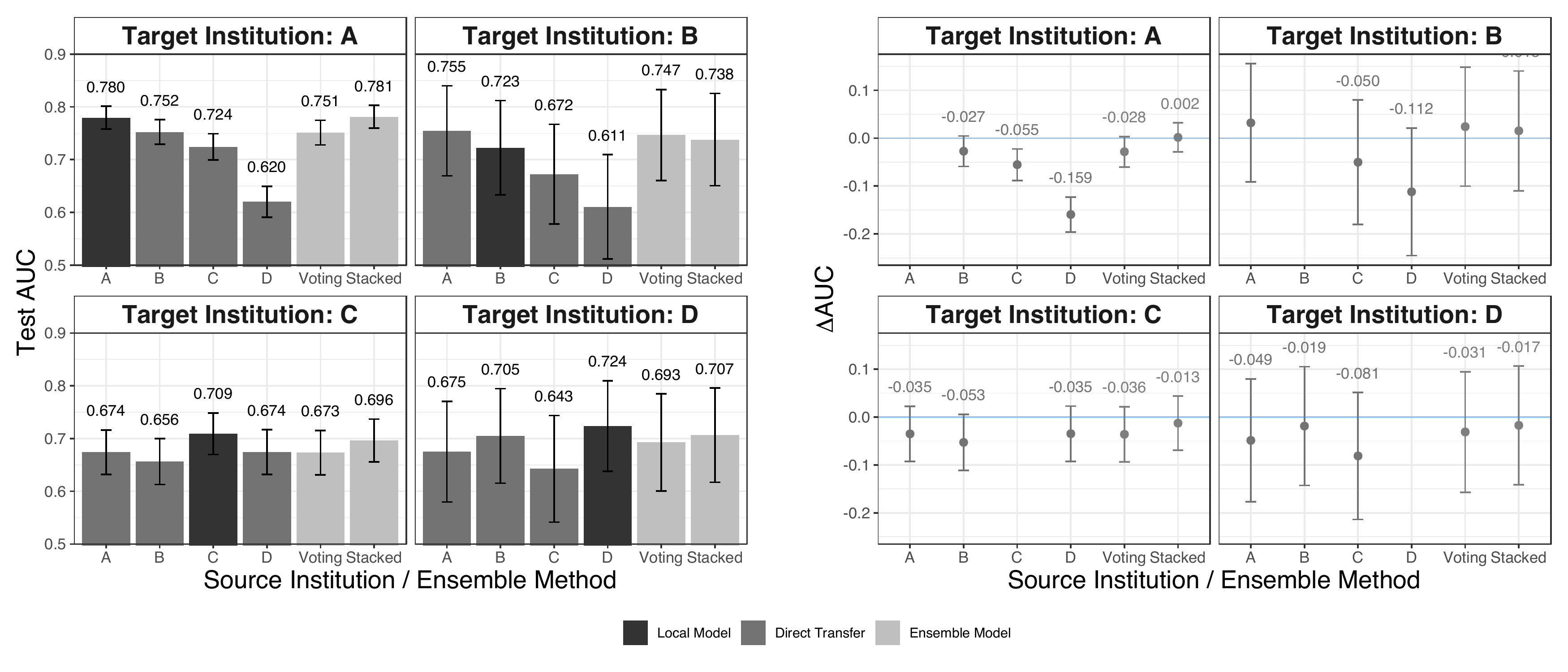}
  \caption{Results for MLP model.}
  \label{fig:auc-dauc-mlp}
\end{subfigure}%
\caption{Additional results for LightGBM (\ref{fig:auc-dauc-lightgbm}) and MLP (\ref{fig:auc-dauc-mlp}) models. The findings are consistent with our findings for $L_2$-regularized logistic regression shown in Figure \ref{fig:auc-dauc-l2lr}: the 95\% CI for AUC for voting transfer overlaps with the local model, for all institutions and for both LightGBM and MLP. Additionally, the 95\% CI for $\dauc$ overlaps with zero, indicating no transfer gap between the voting transfer model and the local model.}
\label{fig:auc-dauc-lightgbm-mlp}
\end{figure*}

\begin{figure*}
\centering
\begin{subfigure}{\textwidth}
  \centering
  \includegraphics[width=\textwidth]{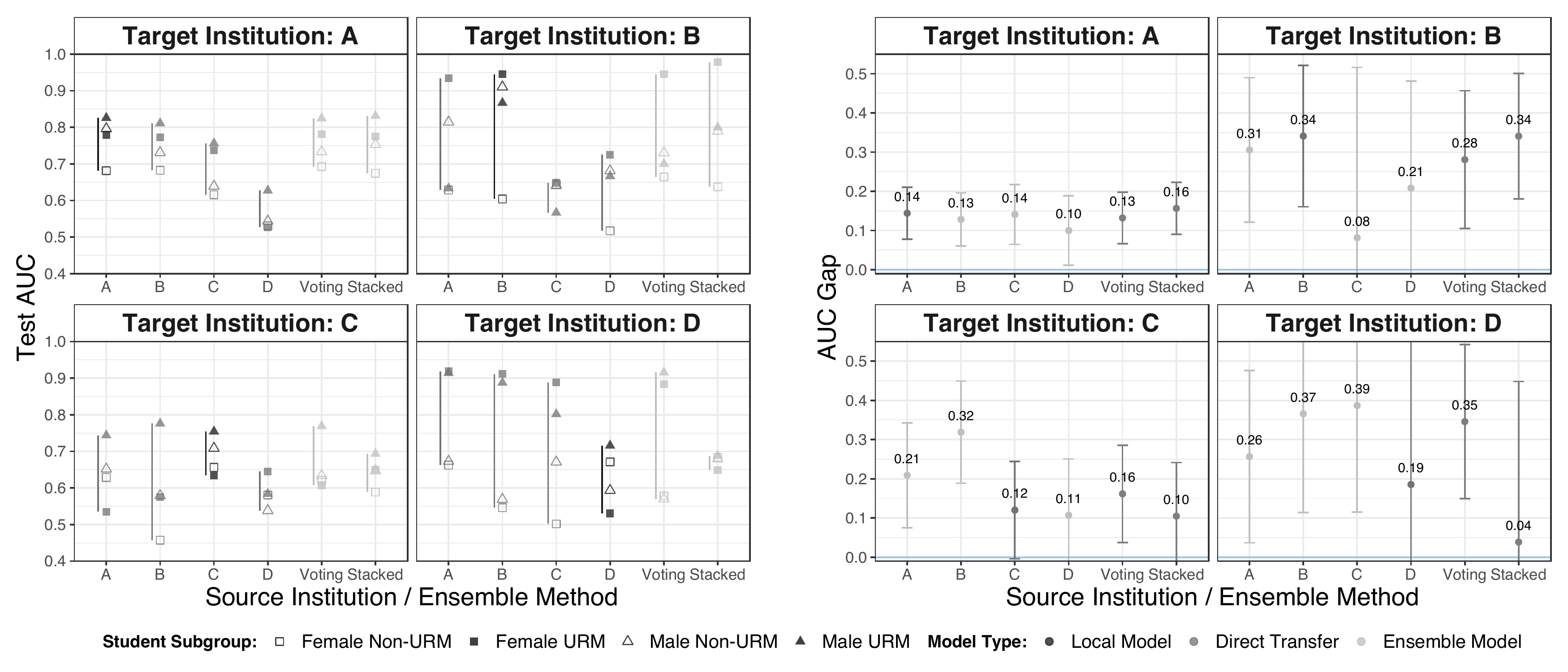}
  \caption{Results for LightGBM model.}
  \label{fig:aucgap-lightgbm}
\end{subfigure}
\begin{subfigure}{\textwidth}
  \centering
  \includegraphics[width=\textwidth]{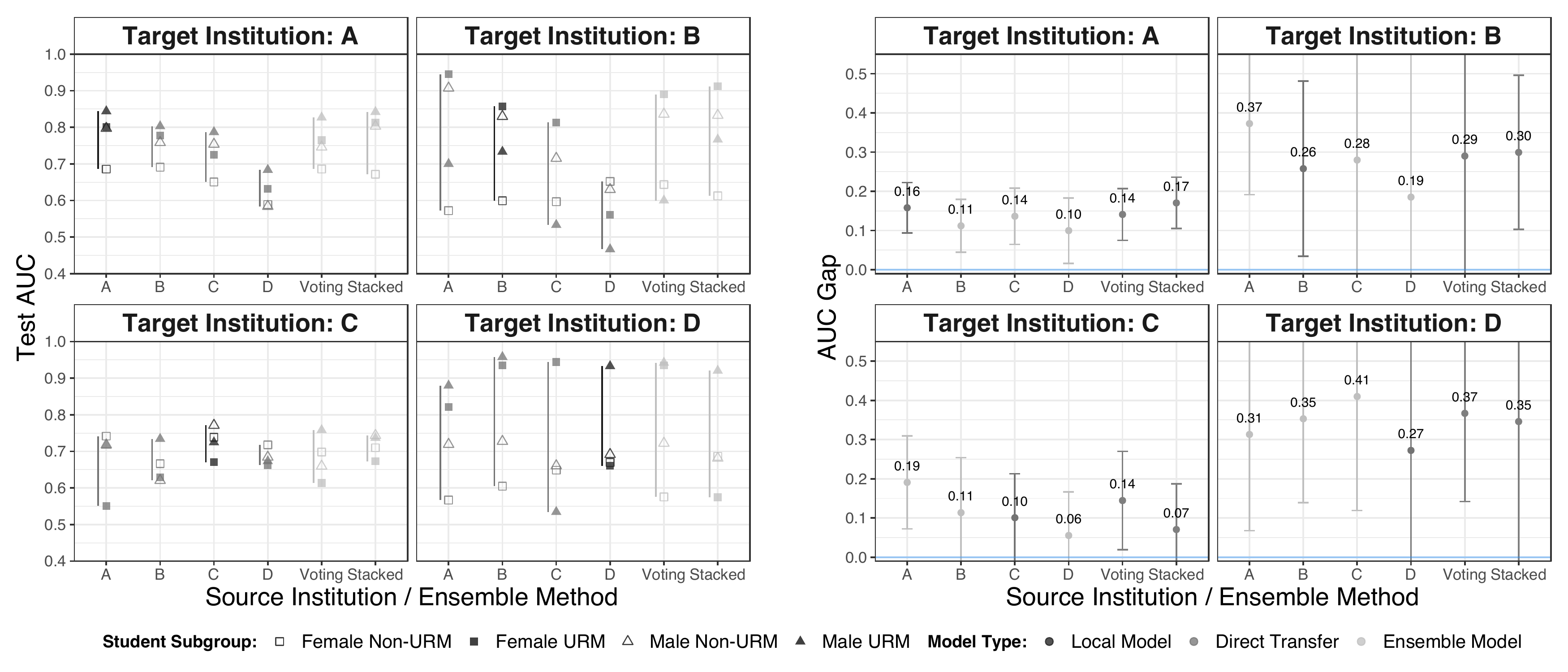}
  \caption{Results for MLP model.}
  \label{fig:aucgap-mlp}
\end{subfigure}%
\caption{Additional results for LightGBM (\ref{fig:aucgap-lightgbm}) and MLP (\ref{fig:aucgap-mlp}) models. The findings are consistent with our findings for $L_2$-regularized logistic regression shown in Figure \ref{fig:aucgap-l2lr}: the 95\% CI for AUC Gap for voting transfer overlaps with the local model, for all institutions and for both LightGBM and MLP, indicating no difference in the intersectional subgroup performance disparities between the voting transfer model and the local model.}
\label{fig:aucgap-lightgbm-mlp}
\end{figure*}

We provide identical experimental results as our figures in the main text here, using LightGBM and MLP models in place of L2LR. Figure \ref{fig:overall-lightgbm-mlp} shows results analogous to main text Figure \ref{fig:overall-l2lr}; Figure \ref{fig:auc-dauc-lightgbm-mlp} shows results analogous to main text Figure \ref{fig:auc-dauc-l2lr}; and Figure \ref{fig:aucgap-lightgbm-mlp} provides results analogous to Figure \ref{fig:aucgap-l2lr}. The results shown for LightGBM and MLP models are consistent with those discussed in the main text.

\subsection{Regularization Study}\label{sec:results-regularization}

\begin{figure*}
    \centering
    \begin{subfigure}{\textwidth}
      \centering
      \includegraphics[width=\textwidth]{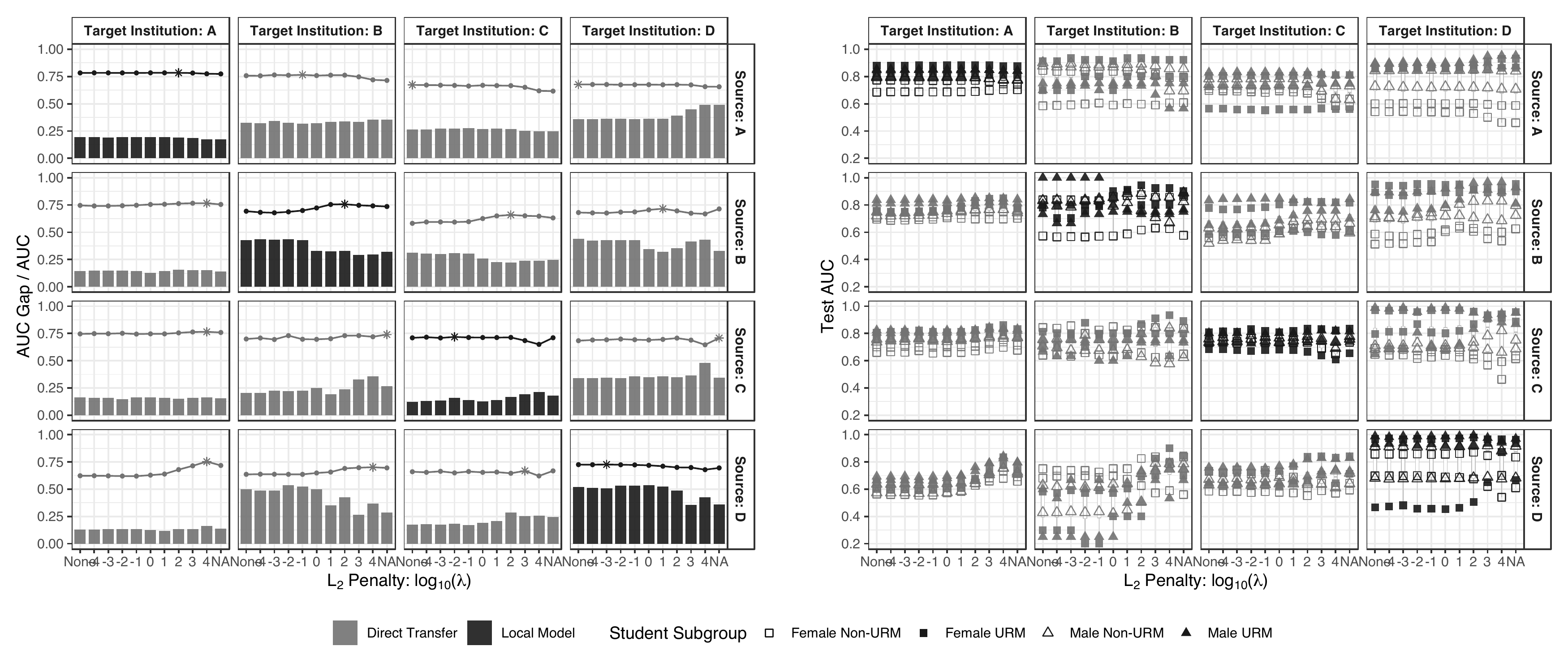}
      \caption{Regularization study results for L2LR model.}
      \label{fig:lambda-study-l2lr}
    \end{subfigure}
    \begin{subfigure}{\textwidth}
      \centering
      \includegraphics[width=\textwidth]{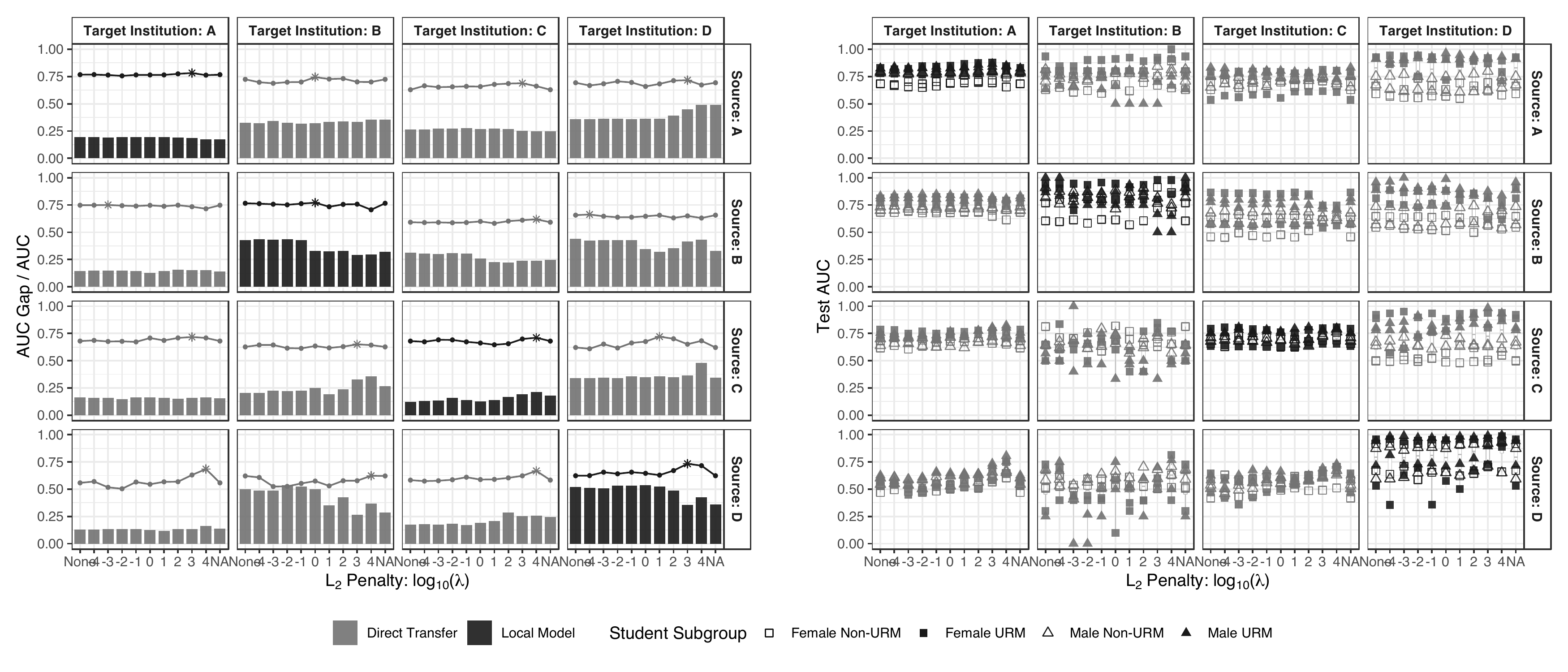}
      \caption{Regularization study results for LightGBM model.}
      \label{fig:lambda-study-lightgbm}
    \end{subfigure}
    \begin{subfigure}{\textwidth}
      \centering
      \includegraphics[width=\textwidth]{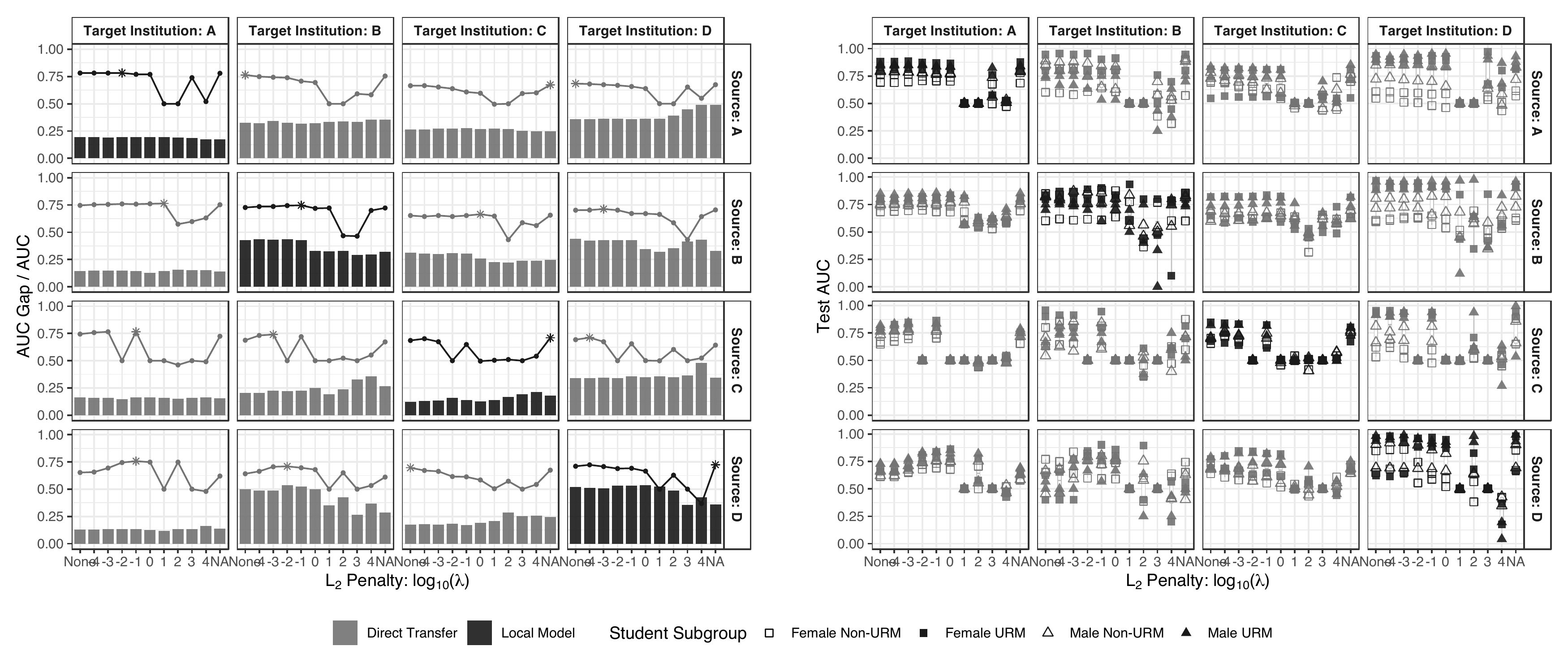}
      \caption{Regularization study results for MLP model.}
      \label{fig:lambda-study-mlp}
    \end{subfigure}

    \caption{Results of regularization study over various $L_2$ regularization strengths for local and direct transfer models.  Left: Test AUC (lines) and AUC Gap (bars). `*' marker indicates the value $\lambda^*$ which achieves highest test AUC for the given model type and source/target institution. Right: Intersectional subgroup model performance.}
    \label{fig:lambda-study}
\end{figure*}

Prior work suggested that \textit{regularization} affects model fairness by controlling worst-group outcomes, including in modeling regimes relevant to our fairly simple $L_2$-regularized logistic regression approach. For example, \cite{kamishima2011fairness} shows that a regularizer consisting of an $L_2$ penalty, combined with a ``prejudice removal'' regularizer, can reduce a measure of unfairness. However, their specific formulation of unfairness seeks to minimize the model's reliance on sensitive attributes to avoid disparate treatment, while in our experiments, we do not seek to explicitly avoid this. Using distributionally-robust neural network training, \cite{sagawa2019distributionally} showed that increasing regularization (via increasing an $L_2$ penalty or via early stopping) improves worst-group accuracy. \cite{khani2020feature} suggested that the removal of spurious (i.e. non-informative) features can have disproportionate effects on subgroups. We are aware of no previous work which explores the effect of regularization on cross-institutional transfer, despite the clear connection between regularization, the bias-variance tradeoff, and generalization error, which could have implications for domain transfer.

In this section, we conduct an exploratory study of the impact of regularization on overall performance, and on equitable performance over intersectional subgroups. Our procedure is as follows: for each training experiment above, we fix the $L_2$ regularization penalty parameter $\lambda \in {0, 10^{-4}, 10^{-3},  \ldots, 10^4}$ and follow the same training and evaluation procedure. We only conduct this for \textit{local} and \textit{direct} transfer scenarios. Our aim in this study is to determine how regularization might affect transfer, and how specific subgroups are affected.

Results of this study are shown in Figure~\ref{fig:lambda-study}. First, the dotted lines in Figure~\ref{fig:lambda-study}a indicate AUC at different levels of regularization, which shows the standard expected result that (due to bias-variance tradeoff) regularization tends to have an ``optimal'' value (indicated by a `*' dot), above or below which a model's test performance tends to decline.

Figure~\ref{fig:lambda-study}b shows that each intersectional subgroup tends to respond similarly to regularization, and that changes in regularization generally fail to reduce performance disparities across all source-target institution pairs, with the rank-ordering of model performance for subgroups largely consistent across source institutions. 

We discuss the results in further detail in Section \ref{sec:results}.

\begin{figure*}
    \centering
    \includegraphics[width=3in]{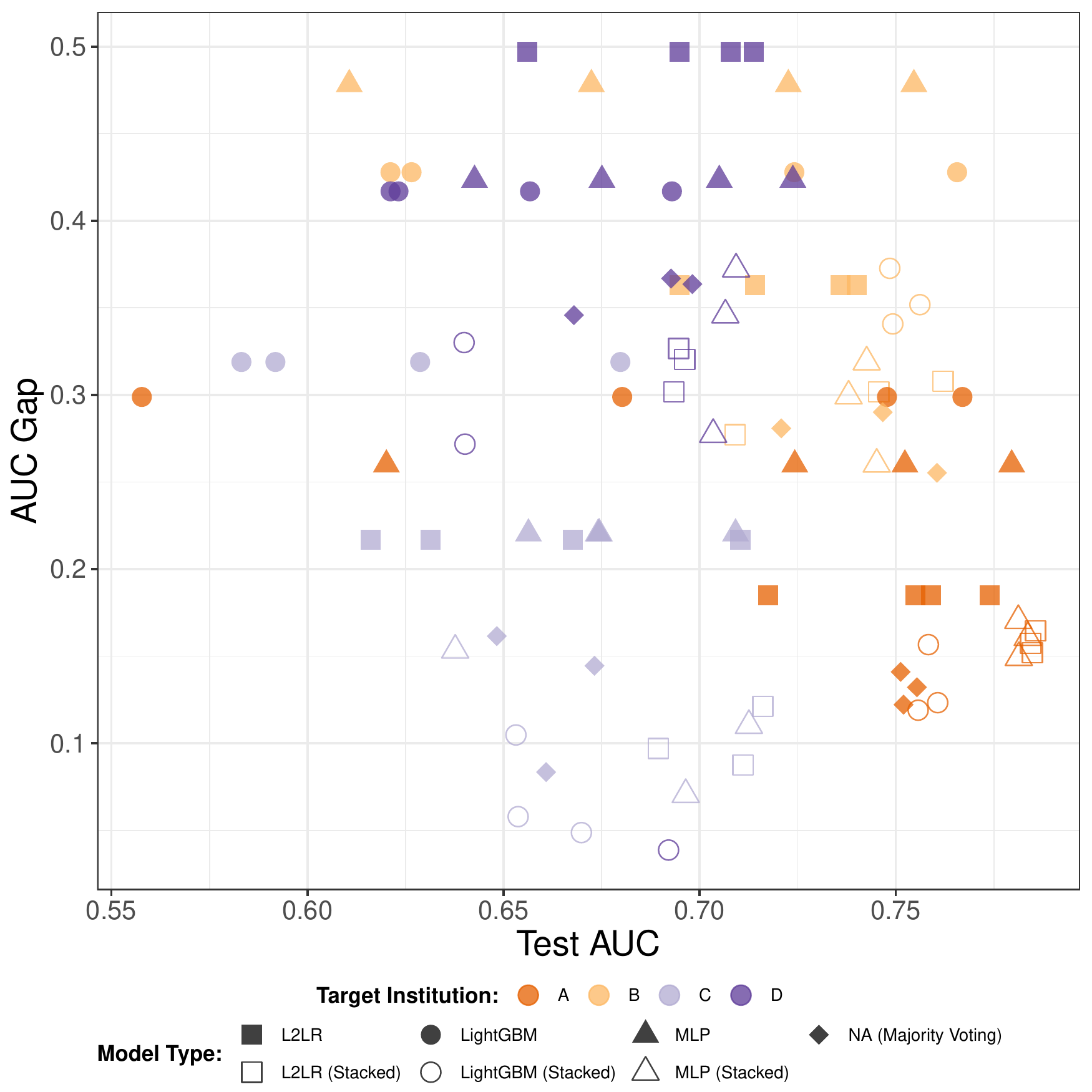}
    \caption{Overall performance (measured by AUC) vs. fairness (AUC Gap, Eq. \eqref{eqn:auc-gap}).}
    \label{fig:auc-vs-aucgap}
\end{figure*}

\subsection{Model Similarity Analysis}\label{sec:model-similarity-analysis}

While our experiments demonstrate that models trained locally at each institution have similar average performance to each other, our experiments do not verify that these models learn similar functions of the inputs; instead, they merely verify that their average performance on each institution's test set is similar. In this section, we briefly explore whether the learned \textit{coefficients} are similar between local models learned at each institution.

A standard method to compare whether logistic regression coefficients differ between two datasets requires multi-institutional training (having access to \textit{all} datasets), and would involve adding an \texttt{institution} indicator variable, and determining whether the coefficient corresponding to this variable was nonzero in the fitted model. However, since we are unable to train directly on multi-institutional datasets due to the privacy constraints mentioned above, we explore alternative methods for model comparison. These methods should be considered qualitative explorations of the similarity of the learned models.

\begin{figure*}
\centering
\begin{subfigure}{.45\textwidth}
  \centering
  \includegraphics[width=.85\textwidth]{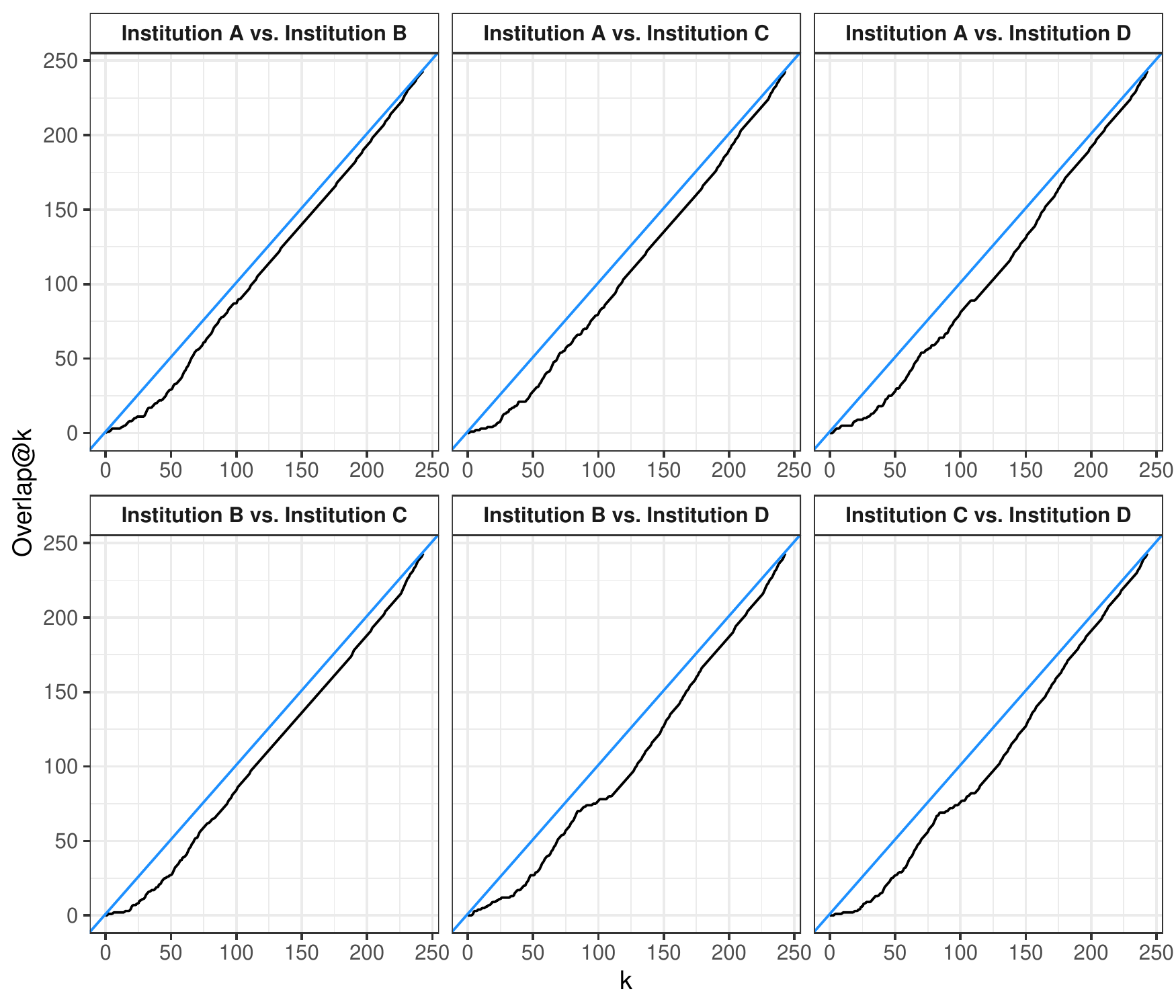}
  \caption{Model similarity metric $\textrm{Overlap@}k$ for each pair of local models.}
  \label{fig:overlap-at-k}
\end{subfigure}
\begin{subfigure}{.45\textwidth}
  \centering
  \includegraphics[width=.85\textwidth]{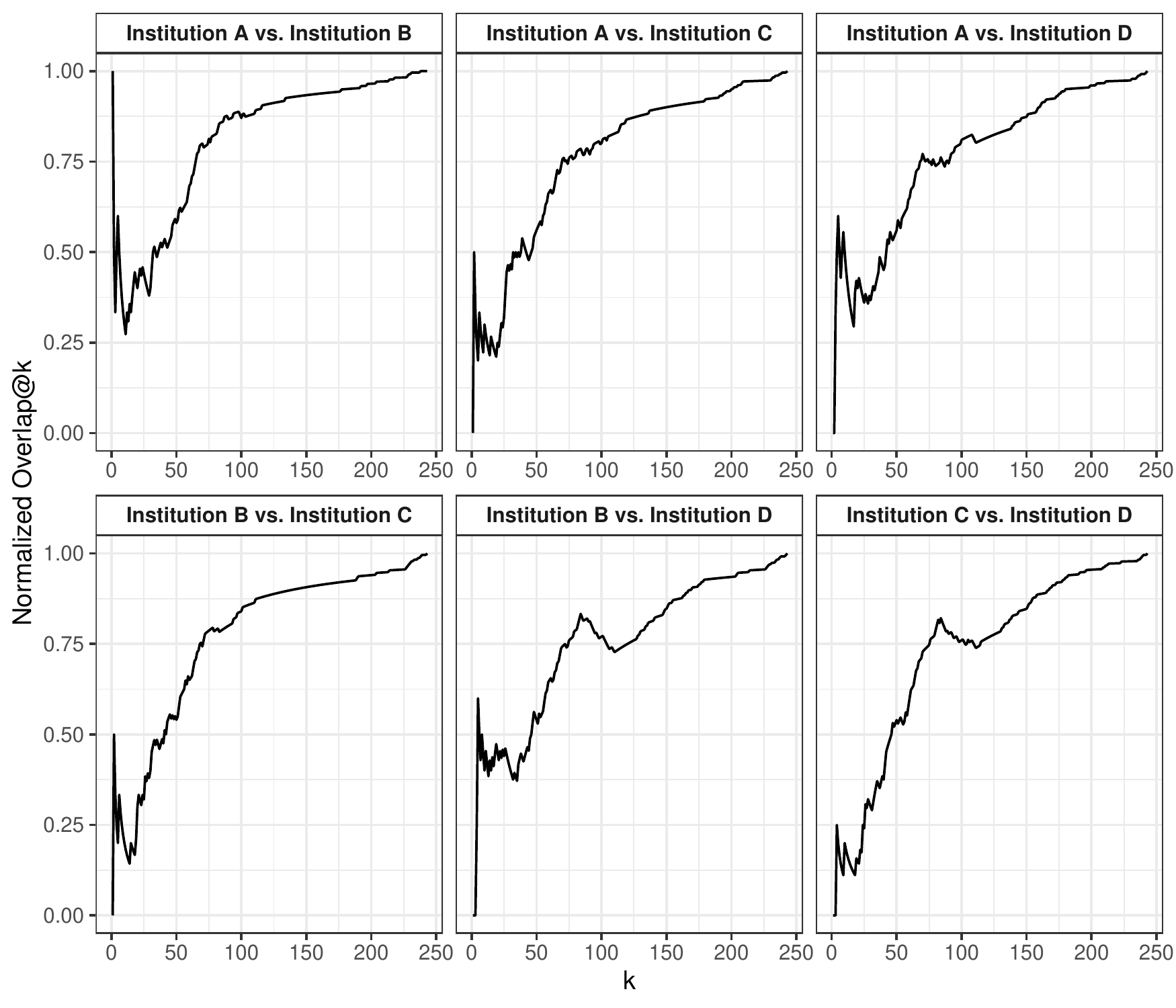}
  \caption{Model similarity metric $\textrm{Normalized~Overlap@}k$ for each pair of local models.}
  \label{fig:overlap-at-k-normalized}
\end{subfigure}%
\caption{The proposed model similarity measures for each pair of locally-learned models.}
\label{fig:overlap}
\end{figure*}

We provide two exploratory analyses to address this question. First, in Figure \ref{fig:overlap}, we compute an overlap metric, $\textrm{Overlap}@k$, for the local models from each pair of institutions. $\textrm{Overlap}@k$ is computed as follows: let $\hat{\theta} = [\theta_1, \ldots, \theta_d]$ represent the $d$-dimensional coefficients of a model. For each pair of $\hat{\theta}_i, \hat{\theta}_j$, we separately sort the elements in descending order by magnitude $sort(\hat{\theta})$. Then, for fixed $k$, we take the first $k$ elements of both vectors and compute the size of the overlap:

\begin{equation}
    \textrm{Overlap}@k(\hat{\theta}_i, \hat{\theta}_j) = \big| sort(\hat{\theta}_i)[1:k] \cap sort(\hat{\theta}_j)[1:] \big|
\end{equation}

Intuitively, $\textrm{Overlap}@k$ measures the level of agreement between models about which coefficients are largest in magnitude. This does not, for example, ensure that these coefficients have even the same direction, but it provides a qualitative measure of agreement on feature importance. Two models which have identical rank-ordering of feature magnitudes would have $\textrm{Overlap}@k$ of $k$, the highest possible value; two models which do not agree on any of the highest-$k$-magnitude features would have $\textrm{Overlap}@k$ of zero.

Because it can be easier to interpret the overlap as the relative size of the intersection, we also report the Normalized $\textrm{Overlap}@k$, obtained by normalizing by a factor of $1/k$:

\begin{equation}
    \textrm{Normalized~Overlap}@k(\hat{\theta}_i, \hat{\theta}_j) = \frac{1}{k} \big| sort(\hat{\theta}_i)[1:k] \cap sort(\hat{\theta}_j)[1:] \big|
\end{equation}

Results of our computation of these similarity metrics are shown in Figure \ref{fig:overlap}.

As an additional check of model similarity, which (unlike $\textrm{Overlap@}k$) accounts for the \textit{directionality} of the feature vectors, we also report the cosine similarity between each pair of model coefficients. Cosine similarity is a measure of the angle between two vectors, irrespective of their magnitudes, and is defined as:

\begin{equation}
  cossim(\hat{\theta}_i, \hat{\theta}_j) = \frac{\langle \hat{\theta}_i, \hat{\theta}_j \rangle}{||\hat{\theta}_i || \hat{\theta}_j||}
\end{equation}

\begin{table}[!h]
\begin{tabular}{c S[round-mode=places, round-precision=2] S[round-mode=places, round-precision=2] S[round-mode=places, round-precision=2]}
\toprule
Institution & {B}          & {C}          & {D}           \\ \midrule
A           & 0.46448893 & 0.397752 & 0.26986666  \\
B           &            & 0.33703219 & 0.2225624  \\
C           &            &            & 0.17134966 \\ \bottomrule
\end{tabular}
\caption{Model similarity measure $cos(\hat{\theta}(X_i), \hat{\theta}(X_{i'})$.}
\end{table}\label{tab:cossim}
\npnoround

Results of this comparison are shown in Table \ref{tab:cossim}

\section{Computing Standard Errors of AUC}\label{section:seauc}

As discussed above, we compute standard errors for AUC estimates according to the procedure described in \citep{hanley1982meaning, fogarty2005case}, which utilizes the equivalence between the Area Under the Receiver Operating Characteristic Curve and the Wilcoxon Statistic.

Formally, let $n_p = \sum_{i=1}^n \mathbbm{1}(y_i = 1)$ and $n_n = \sum_{i=1}^n \mathbbm{1}(y_i = 0)$ be defined as the number of positive and negative examples in the dataset of interest, respectively. Define $A'$ as the AUC on the dataset. Then, let

\begin{align}
    D_p &\coloneqq (n_p - 1) (\frac{A'}{2 - A'} - A'^2) \\
    D_n &\coloneqq (n_n - 1) (\frac{2 A'^2}{1 + A'} - A'^2)
\end{align}

Then the standard error of $A'$ can be computed as:

\begin{equation}
    SE(A') = \sqrt{\frac{A' (1 - A') + D_p + D_n}{n_p n_n}}\label{eqn:seauc}
\end{equation}

For further details and proof, we defer the reader to \cite{hanley1982meaning}.

\end{document}